\documentclass[preprint,12pt]{elsarticle}
\usepackage{lipsum}
\usepackage{lineno}
\usepackage{natbib}
\usepackage{pdflscape}
\usepackage{bm,multicol,color}
\usepackage{amsmath,lscape}
\usepackage{mathtools}
\usepackage{amssymb}
\usepackage[utf8]{inputenc}
\usepackage{epsfig}
\usepackage{amsfonts}
\usepackage{fullpage}
\usepackage[flushleft]{threeparttable}
\usepackage{rotating,array}
\usepackage{float}
\usepackage{comment}
\usepackage{fvextra}  
\usepackage{graphicx}
\usepackage{makecell}
\usepackage{multirow}
\usepackage{siunitx}
\usepackage{lscape}
\usepackage{listings}
\usepackage[nolist]{acronym}
\usepackage{longtable}
\usepackage{pdflscape}
\usepackage{afterpage}
\usepackage{slashbox}
\usepackage{wrapfig}
\usepackage{algorithm}
\usepackage{algpseudocode}
\usepackage{enumitem}
\usepackage{systeme}
\usepackage[breaklinks=true]{hyperref}
\hypersetup{
    unicode=true,          
    pdftoolbar=true,        
    pdfmenubar=true,        
    pdffitwindow=false,     
    pdfstartview={FitH},    
    pdfnewwindow=true,      
    colorlinks=true,        
    linkcolor= black,        
    citecolor= black,        
    filecolor= black,        
    urlcolor= black          
}
\usepackage{array}
    \newcolumntype{C}[1]{>{\centering\arraybackslash}m{#1}}
\usepackage{color,soul}
\usepackage{graphicx}
\usepackage{subcaption}
\usepackage{booktabs}
\usepackage{fullpage}
\usepackage[vlined,ruled,algo2e]{algorithm2e}
\usepackage{svg}
\usepackage{graphics}
\usepackage{listings}
\usepackage{xcolor}
\usepackage{tikz}
\usetikzlibrary{positioning, shapes.geometric, arrows.meta}
\definecolor{codegreen}{rgb}{0,0.6,0}
\definecolor{codegray}{rgb}{0.5,0.5,0.5}
\definecolor{codepurple}{rgb}{0.58,0,0.82}
\definecolor{backcolour}{rgb}{0.95,0.95,0.92}

\lstdefinestyle{mystyle}{
    backgroundcolor=\color{backcolour},   
    commentstyle=\color{codegreen},
    keywordstyle=\color{magenta},
    numberstyle=\tiny\color{codegray},
    stringstyle=\color{codepurple},
    basicstyle=\ttfamily\footnotesize,
    breakatwhitespace=false,         
    breaklines=true,                 
    captionpos=b,                    
    keepspaces=true,                 
    numbers=left,                    
    numbersep=5pt,                  
    showspaces=false,                
    showstringspaces=false,
    showtabs=false,                  
    tabsize=2
}
\begin{document}

\begin{frontmatter}
        \title{A novel YOLO26-MoE optimized by an LLM agent for insulator fault detection considering UAV images}

\author[LASIGE]{Jo\~ao Pedro Matos-Carvalho}
\ead{jpecarvalho@ciencias.ulisboa.pt}
\author[autUFSC]{Laio Oriel Seman}
\ead{laio@gos.ufsc.br}
\author[ISEL]{Stefano Frizzo Stefenon}
\ead{stefano.stefenon@isel.pt}
\author[SAL]{Mohammad Khalaf Mohammad Khreasat} 
\ead{idu078144@usal.es}
\author[SAL]{Gabriel Villarrubia Gonz\'alez}
\ead{gvg@usal.es}

\address[LASIGE]{LASIGE, Faculdade de Ciências, Universidade de Lisboa, 1749-016 Lisboa, Portugal}

\address[autUFSC]{Department of 
Automation and Systems 
Engineering, Federal University of Santa Catarina, Florianópolis, Brazil}

\address[ISEL]{Instituto Superior de Engenharia de Lisboa, Instituto Politécnico de Lisboa, Rua Conselheiro Emídio Navarro 1, 1959-007 Lisboa, Portugal}

\address[SAL]{Expert Systems and Applications Lab, Faculty of Science, University of Salamanca, Plaza de los Caídos s/n, 37008 Salamanca, Spain}

\journal{Elsevier}

\begin{abstract}
The inspection of electrical power line insulators is essential for ensuring grid reliability and preventing failures caused by damaged or degraded insulation components. In recent years, Unmanned Aerial Vehicles (UAVs) combined with deep learning-based vision systems have emerged as an effective solution for automating this process. However, insulator fault detection remains challenging due to small defect regions, heterogeneous fault patterns, complex backgrounds, and varying imaging conditions. To address these challenges, this paper proposes an optimized YOLO26-MoE, a novel object detection architecture that integrates a sparse Mixture-of-Experts (MoE) module into the high-resolution branch of the YOLO26 detector. The proposed modification enables adaptive feature refinement for subtle and diverse fault patterns while preserving the efficiency of a one-stage detection framework. Hyperparameter optimization, final training, and evaluation were coordinated through a tool-augmented Large Language Model (LLM) agent. The proposed model achieved 0.9900 mAP@0.5 and 0.9515 mAP@0.5:0.95, outperforming the latest YOLO versions. These results demonstrate that the proposed model provides an effective and reliable solution for UAV-based insulator fault detection.
\end{abstract}

\begin{keyword}
Insulator fault detection \sep UAV inspection \sep YOLO26 \sep Mixture-of-Experts \sep Large language model.
\end{keyword}

\end{frontmatter}



\section{Introduction}

The reliable operation of electrical power systems depends on the condition of their insulation components. Insulators are responsible for maintaining electrical isolation between conductors and grounded structures, preventing current leakage and ensuring the safety and efficiency of power transmission \cite{10478197}. However, insulators are continuously exposed to environmental stressors such as pollution, ultraviolet radiation, humidity, and mechanical load, which can lead to surface degradation, cracking, or flashover \cite{STEFENON2025110682}. These faults may result in partial discharges, power outages, and even large-scale blackouts if not detected in time \cite{10534789}. Consequently, accurate and timely detection of faulty insulators is a vital task for maintaining grid stability and minimizing maintenance costs.

Traditional inspection, including visual assessment and manual thermographic or ultraviolet imaging, is often labor-intensive, subjective, and limited in scalability. In recent years, deep learning has emerged as a powerful approach to automate and enhance fault detection in insulators \cite{corso2023evaluation}. Through convolutional and transformer-based neural networks, deep learning models can extract complex spatial and spectral features from infrared, ultraviolet, and visible imagery, enabling high-precision classification and localization of defects. These models reduce human intervention, facilitating the development of intelligent condition-monitoring systems for power grids \cite{10750545}.

The deployment of Unmanned Aerial Vehicles (UAVs) has further revolutionized power line inspection by enabling rapid data acquisition over vast and often inaccessible terrain \cite{zhang2022uav}. UAV-mounted cameras can capture high-resolution images of insulators from multiple angles and distances, providing rich visual information for automated analysis. However, UAV-acquired imagery presents unique challenges, including varying illumination conditions, complex backgrounds, motion blur, and diverse insulator orientations, which demand robust and adaptive detection algorithms \cite{liu2021review}.

Among deep learning architectures, the You Only Look Once (YOLO) family of detectors has gained widespread adoption for real-time object detection tasks due to its favorable trade-off between accuracy and computational efficiency \cite{redmon2016you}. Successive iterations of YOLO have introduced architectural innovations such as feature pyramid networks, attention mechanisms, and anchor-free detection heads, progressively improving detection performance across diverse domains. Recent versions, including YOLOv8 and beyond, have demonstrated state-of-the-art results on general object detection benchmarks, yet their application to specialized tasks such as insulator fault detection remains an active area of research \cite{chen2022lightweight}. 

More broadly, automated power-line inspection has been increasingly supported by UAV-based vision, robotic inspection platforms, and intelligent sensing pipelines. Prior studies have highlighted the growing relevance of deep learning as the analytical backbone of automatic inspection, while also emphasizing the operational importance of scalable data acquisition and condition assessment in complex transmission environments \citep{nguyen2018vision_powerline,alhassan2020inspection_robots,manninen2021condition_assessment,guan2021uav_lidar_powerline}.

Despite these advances, existing approaches face several limitations. First, the heterogeneity of insulator types, fault categories, and imaging conditions poses challenges for models trained on limited datasets. Second, the long-tailed distribution of fault types, where certain defects occur far more frequently than others, can bias model predictions toward common classes while neglecting rare but critical faults. Third, the manual tuning of hyperparameters and architectural choices remains a time-consuming process that requires substantial domain expertise.

Mixture-of-Experts (MoE) architectures offer a promising paradigm to address the first two challenges \cite{shazeer2017outrageously}. By routing inputs to specialized expert subnetworks based on learned gating mechanisms, MoE models can develop distinct processing pathways for different input characteristics, such as fault types or insulator appearances, without proportional increases in computational cost. This conditional computation enables greater model capacity and specialization while maintaining inference efficiency suitable for real-time applications.

To address the challenge of hyperparameter optimization, recent research has explored the use of Large Language Models (LLMs) as intelligent optimization agents \cite{yang2024large}. Unlike traditional optimization algorithms that operate purely on numerical representations, LLM-based agents can leverage pre-trained knowledge about neural network architectures, interpret experimental results through natural language reasoning, and adaptively refine search strategies based on observed patterns \cite{huang2024large}. This emerging paradigm offers the potential to accelerate the optimization process while incorporating domain-specific insights that would otherwise require extensive human expertise.

This work proposes an optimized YOLO26-MoE, a novel architecture that integrates MoE layers into a state-of-the-art YOLO26 backbone for insulator fault detection in UAV imagery. The model is optimized through an LLM-based agent that leverages domain knowledge about insulator characteristics, fault types, and UAV imaging conditions to guide the hyperparameter search process. The main contributions of this paper are as follows:

\begin{itemize}
    \item A novel YOLO26-MoE architecture that incorporates sparse MoE layers into the detection backbone, enabling specialized processing pathways for diverse insulator types and fault categories while maintaining computational efficiency.
    \item An LLM-based optimization agent that combines natural language reasoning with systematic hyperparameter search, leveraging pre-trained knowledge about computer vision architectures to accelerate convergence and improve final model performance.
    \item Comprehensive experimental evaluation on a UAV-acquired insulator dataset, demonstrating that the proposed approach achieves superior detection accuracy compared to baseline YOLO variants and existing insulator detection methods.
    \item Ablation studies analyzing the contribution of MoE components and the effectiveness of LLM-guided optimization, providing insights into the design choices that drive performance improvements.
\end{itemize}

The remainder of this paper is organized as follows: Section \ref{sec2} reviews related work on insulator defect detection, YOLO based detectors, Mixture of Experts models, and LLM-driven optimization, establishing the research gap addressed in this work. Section \ref{sec3} presents the technical background of YOLO26, including its architectural design and training strategies. Section \ref{sec4} describes the proposed YOLO26 MoE methodology, detailing the integration of the sparse Mixture of Experts module and the LLM agent responsible for hypertuning, training, and evaluation. Section \ref{sec5} presents the experimental setup, dataset, hardware configuration, evaluation metrics, and benchmarking results. Finally, Section \ref{sec6} concludes the paper by summarizing the main findings and outlining future research directions.

\section{Related Work}
\label{sec2}

This section reviews the existing literature on insulator fault detection, object detection architectures for power line inspection, MoE models, and the emerging role of LLMs in hyperparameter optimization.

\subsection{Insulator Fault Detection in Power Systems}

Insulators are critical components in electrical transmission and distribution systems, responsible for mechanically supporting conductors while electrically isolating them from grounded structures. Faults such as contamination, cracks, flashover damage, and missing caps can lead to power outages, equipment damage, and safety hazards~\citep{tao2020detection}.

Traditional inspection methods rely on manual visual inspection by trained personnel, which is time-consuming, labor-intensive, and potentially dangerous, particularly for high-voltage transmission lines in remote or difficult terrain~\citep{liu2021review}. The adoption of UAVs equipped with high-resolution cameras has revolutionized power line inspection by enabling rapid, safe, and cost-effective data acquisition~\citep{zhang2022uav}.

Early automated approaches employed classical image processing techniques, including edge detection, morphological operations, and template matching~\citep{wu2016insulator}. However, these methods struggle with varying lighting conditions, complex backgrounds, and diverse insulator types. The emergence of deep learning has substantially improved detection accuracy and robustness~\citep{zhao2020insulator}. More specifically, recent studies have investigated intelligent recognition strategies for insulator and transmission-line defect analysis under practical field conditions. These works address challenges such as cluttered backgrounds, small defect regions, and limited availability of representative training samples, showing that modern learning-based methods can substantially improve both fault discrimination and localization robustness in realistic inspection scenarios \cite{deng2022edge_intelligent_insulator,liu2022key_target_defect,song2024fault_detection_tlc}.

\subsection{Deep Learning for Object Detection in Power Line Inspection}

Convolutional Neural Networks (CNNs) have become the dominant paradigm for visual inspection tasks. Two-stage detectors, such as faster region-based CNN~\citep{ren2015faster}, achieve high accuracy but suffer from computational overhead that limits real-time deployment. Single-stage detectors, particularly the YOLO family~\citep{redmon2016you}, offer an attractive balance between speed and accuracy for practical applications.

The YOLO architecture has evolved significantly since its introduction. YOLOv3  introduced multi-scale predictions using feature pyramid networks \citep{yang2022bidirection}. YOLOv4~\citep{han2021insulator} incorporated bag-of-freebies and bag-of-specials techniques to improve training and inference. YOLOv5 popularized the architecture with an accessible implementation and efficient training pipeline. More recent versions, including YOLOv7~\citep{wang2023yolov7}, YOLOv8, and subsequent iterations, have continued to push the boundaries of detection performance through architectural innovations such as extended efficient layer aggregation networks and anchor-free detection heads.

Several studies have applied YOLO variants to insulator fault detection. \citet{liu2020insulator} employed YOLOv3 with attention mechanisms to detect insulator defects, achieving improved localization accuracy. \citet{wang2021insulator} proposed a modified YOLOv4 with deformable convolutions to handle the geometric variations of insulators captured from different UAV viewpoints. \citet{chen2022lightweight} developed a lightweight YOLO variant optimized for edge deployment on UAV platforms with limited computational resources.

Despite these advances, existing approaches face challenges in handling the long-tailed distribution of fault types, where certain defects (e.g., missing caps) are significantly more common than others (e.g., flashover marks). Additionally, the trade-off between model complexity and inference speed remains a critical consideration for real-time UAV-based inspection systems.

\subsection{Mixture-of-Experts Architectures}

Mixture-of-Experts, in short MoE, models~\citep{seman2026sparse} represent a paradigm for scaling neural networks by conditionally activating subsets of parameters based on input characteristics. Rather than processing all inputs through the entire network, MoE architectures employ a gating mechanism to route inputs to specialized expert subnetworks, enabling increased model capacity without proportional increases in computational cost.

The seminal work of \citet{shazeer2017outrageously} demonstrated the effectiveness of sparsely-gated MoE layers in scaling language models to unprecedented sizes while maintaining computational efficiency. This approach has since been adopted in various domains, including computer vision~\citep{riquelme2021scaling} and multimodal learning~\citep{mustafa2022multimodal}.

In the context of object detection, MoE architectures offer several potential advantages. Different experts can specialize in detecting objects of varying scales, aspect ratios, or semantic categories~\citep{gan2025mixture}. For insulator fault detection, this specialization is particularly relevant given the diversity of fault types and the varying appearance of insulators across different manufacturers, voltage classes, and environmental conditions.


\subsection{Hyperparameter Optimization and Neural Architecture Search}

The performance of deep learning models is highly sensitive to hyperparameter choices, including learning rate, batch size, data augmentation strategies, and architectural parameters~\citep{bergstra2012random}. Traditional hyperparameter optimization methods include grid search, random search~\citep{bergstra2012random}, and Bayesian optimization~\citep{snoek2012practical}.


Neural Architecture Search (NAS) extends hyperparameter optimization to the architectural level, automatically discovering optimal network topologies~\citep{zoph2017neural}. Efficient NAS methods, including differentiable approaches~\citep{liu2019darts} and weight-sharing strategies~\citep{pham2018efficient}, have reduced the computational cost of architecture search from thousands of GPU-days to more practical timeframes.

For YOLO-based detectors, several studies have employed automated optimization techniques. \citet{wang2021scaled} used NAS to discover efficient scaling strategies for YOLOv4. These approaches typically require substantial computational resources and domain expertise to configure the search space appropriately.

\subsection{Large Language Models as Optimization Agents}

Large language models, in short LLMs, have demonstrated remarkable capabilities in reasoning, code generation, and task planning~\citep{brown2020language}. Recent research has explored leveraging these capabilities for scientific discovery and optimization tasks~\citep{romera2024mathematical}.

The concept of LLM-based optimization agents represents an emerging paradigm where language models guide the search process through natural language reasoning and code generation~\citep{yang2024large}. Unlike traditional optimization algorithms that operate on numerical representations, LLM agents can incorporate domain knowledge, interpret experimental results, and adaptively modify search strategies based on observed patterns.

\citet{chen2024evoprompting} proposed EvoPrompting, which uses LLMs to generate and evolve neural network architectures through natural language descriptions. \citet{liu2025largelanguagemodelagent} demonstrated that Large Language Models can effectively support hyperparameter optimization by iteratively proposing and refining parameter configurations based on observed performance. \citet{jiang2024llmopt} introduced LLMOPT, a framework that integrates LLM-based reasoning with traditional optimization algorithms to enhance sample efficiency.

The application of LLM agents to optimize object detection models for specific domains presents several advantages. First, LLMs can leverage their pre-trained knowledge about computer vision architectures and training practices. Second, they can interpret qualitative feedback about model behavior, such as failure modes on specific object categories. Third, they can generate human-readable explanations for optimization decisions, improving transparency and trust in automated systems.

However, challenges remain in ensuring the reliability and consistency of LLM-based optimization. Hallucination, where models generate plausible but incorrect suggestions, poses risks in safety-critical applications~\citep{ji2023survey}. Additionally, the computational cost of querying LLMs must be balanced against the efficiency gains from improved optimization.

\subsection{Research Gaps and Contributions}

Although prior studies have reported substantial progress in defect detection and inspection automation for power-system assets, most existing approaches still rely on conventional backbone refinements or incremental adaptations of standard object detectors. Comparatively less attention has been given to conditional-computation mechanisms capable of specializing feature processing for heterogeneous visual patterns in power-line inspection scenarios, particularly when subtle defects occupy limited spatial regions and exhibit strong intra-class variability \citep{zhang2025efenet,wang2025yolo_substation}.

Despite significant progress in insulator fault detection and deep learning optimization, several gaps remain in the literature:

\begin{enumerate}[label=(\roman*)]
    \item Existing YOLO variants for insulator detection do not exploit the potential of MoE architectures to handle the heterogeneity of fault types and insulator appearances.
    
    \item The application of state-of-the-art YOLO architectures (beyond YOLOv8) to power line inspection has not been thoroughly investigated.
    
    \item LLM-based optimization agents have not been applied to domain-specific object detection tasks such as insulator fault detection.
    
    \item The integration of MoE components with modern YOLO architectures and their optimization through LLM agents remains unexplored.
\end{enumerate}

This work addresses these gaps by proposing YOLO26-MoE, a novel architecture that incorporates MoE layers into a YOLO26 backbone, optimized through an LLM agent that leverages domain knowledge about insulator characteristics and UAV imaging conditions. The proposed approach aims to achieve superior detection performance while maintaining computational efficiency suitable for practical deployment scenarios.

\section{Background of YOLO26 against contemporary detectors}
\label{sec3}

YOLO26 is an Ultralytics release focused on edge-optimized real-time detection. Its principal design choices are the removal of the distribution focal loss module and native end-to-end inference without non-maximum suppression \cite{sapkota2025YOLO26}. YOLO26 further introduces ProgLoss for progressive balancing of loss terms, Small Target Adaptive Labeling (STAL) for small target label assignment, and Multi-Target Sampling Gradient Descent (MuSGD) as a hybrid optimizer. These design choices target deployment simplicity, quantization robustness, and improved small object performance as described in the Ultralytics report. 

\subsection{Formal Metrics and Trade Off}
Detection systems are commonly evaluated by mean Average Precision (mAP) and inference latency. Let $T$ denote latency in milliseconds per image measured on a target device. A concise way to express the accuracy latency trade-off is by the scalar score
\begin{align}
S_{\alpha} &= \text{mAP} - \alpha \cdot \frac{T}{T_{0}},
\end{align}
where $\alpha$ is a user-chosen penalty factor that encodes how much latency is penalized relative to accuracy, and $T_{0}$ is a reference latency usually set to 1 ms or to the latency of a baseline model. Models optimal for edge use will maximize $S_{\alpha}$ for a relatively large $\alpha$ while server-oriented models tolerate a smaller $\alpha$.

\subsection{Architectural and Training Differences}

YOLOv8 uses a decoupled head anchor-free design and, in practice, benefited from Distribution Focal Loss for fine bounding box localization and from standard post-processing based on non-maximum suppression \cite{10298815}. These elements provided high accuracy on server-class Graphics Processing Units (GPUs) while maintaining promising speed under optimized runtimes.

YOLOv9 emphasized optimization of internal gradient flow and efficient feature reuse, considering Generalized Efficient Layer Aggregation Network (G-ELAN) \cite{10854439}, while YOLOv10 advanced toward fully end-to-end detection with Optimal Transport-based label assignment and elimination of post-processing \cite{10930899}.

YOLOv11 introduced smaller CSP kernel blocks for efficiency and spatial attention modules to guide focus toward salient regions \cite{11053860}. YOLOv12 emphasized attention-centric modules to capture global context \cite{11113267}. 
These models improved raw accuracy, particularly on large, complex datasets at the cost of increased architectural complexity and greater sensitivity to quantization \cite{11218830}.

Detectors that rely on transformer encoders and decoders emphasize end-to-end training and global attention at the architectural level. Those models often yield strong, large object accuracy and improved contextual reasoning \cite{10669376}. However, these models are typically more sensitive to quantization and incur higher latency on Central Processing Unit (CPU) and some edge accelerators.

YOLO26 departs from the recent complexity trend by removing the distribution focal loss module and designing a native end-to-end predictor that does not require non-maximum suppression. Training stability and small object recovery are improved by ProgLoss and STAL, respectively. The MuSGD optimizer accelerates convergence and reduces hyperparameter fragility \cite{sapkota2025YOLO26}. The combined effect is a model with improved export friendliness, consistent quantization performance, and notably lower CPU latency in nano-scale variants. Empirical claims and deployment guidance are reported in the Ultralytics technical manuscript.

\begin{table}[ht]
\centering
\caption{Selected qualitative comparison of detectors}
\begin{tabular}{p{0.1\textwidth} p{0.3\textwidth} p{0.5\textwidth}}
\toprule
Model & Key innovations & Deployment implications \\
\midrule
YOLOv8 & Decoupled head anchor free predictions DFL  & High accuracy on GPU servers exportable to common runtimes, moderate complexity. \\

YOLOv9 & Programmable Gradient Information G-ELAN & Improved gradient flow and accuracy on medium and large-scale datasets maintain real-time inference. \\
YOLOv10 & End-to-end training removes non-maximum suppression & Simpler inference pipeline, stable recall, and better global optimization efficiency. \\

YOLO11 & Efficient CSP variants spatial attention & Improved accuracy with modest increase in latency and export complexity. \\
YOLO12 & Attention-centric modules for global context & Stronger large object accuracy increased sensitivity to quantization. \\
YOLO26 & NMS free end-to-end inference removal of DFL ProgLoss, STAL, MuSGD & Lower CPU latency improved quantization tolerance simpler export to ONNX TensorRT CoreML TFLite and OpenVINO empirical gains for small object detection and edge scenarios.\\
\bottomrule
\end{tabular}
\end{table}

\subsection{Practical Guidance}
\begin{enumerate}
\item When the target device is a CPU or low-power accelerator, prefer YOLO26 for real-time inference and for stable quantized performance.
\item When absolute top line accuracy on large server-class GPUs is the objective, consider attention-enhanced versions such as YOLOv12 or transformer detectors, remembering that increased latency and quantization sensitivity are likely.
\item When small object recall matters, use training strategies similar to ProgLoss and STAL or adopt multi-scale high-resolution training. YOLO26 implements these ideas natively and shows improved small object performance in reported benchmarks.
\end{enumerate}

YOLO26 embodies a design shift that privileges deployment simplicity, quantization robustness, and deterministic inference behavior while retaining competitive accuracy. For studies where edge deployment, real-time throughput, and small object robustness are priorities, YOLO26 is the recommended baseline. For studies oriented toward large-scale server benchmarks and research into global context modeling, transformer-based detectors remain strong alternatives.

\section{Methodology}
\label{sec4}

In this section, we describe the proposed methodology adopted in this work. The object detector used in this paper is a modified version of YOLO26, hereafter referred to as YOLO26-MoE, in which a sparse MoE module is introduced into the small-object detection branch of the head. Hyperparameter optimization, final training, evaluation, and qualitative inspection are coordinated by a tool-augmented LLM agent.

\begin{figure}[!ht]
    \centering    
    \includegraphics[width=0.99\linewidth]{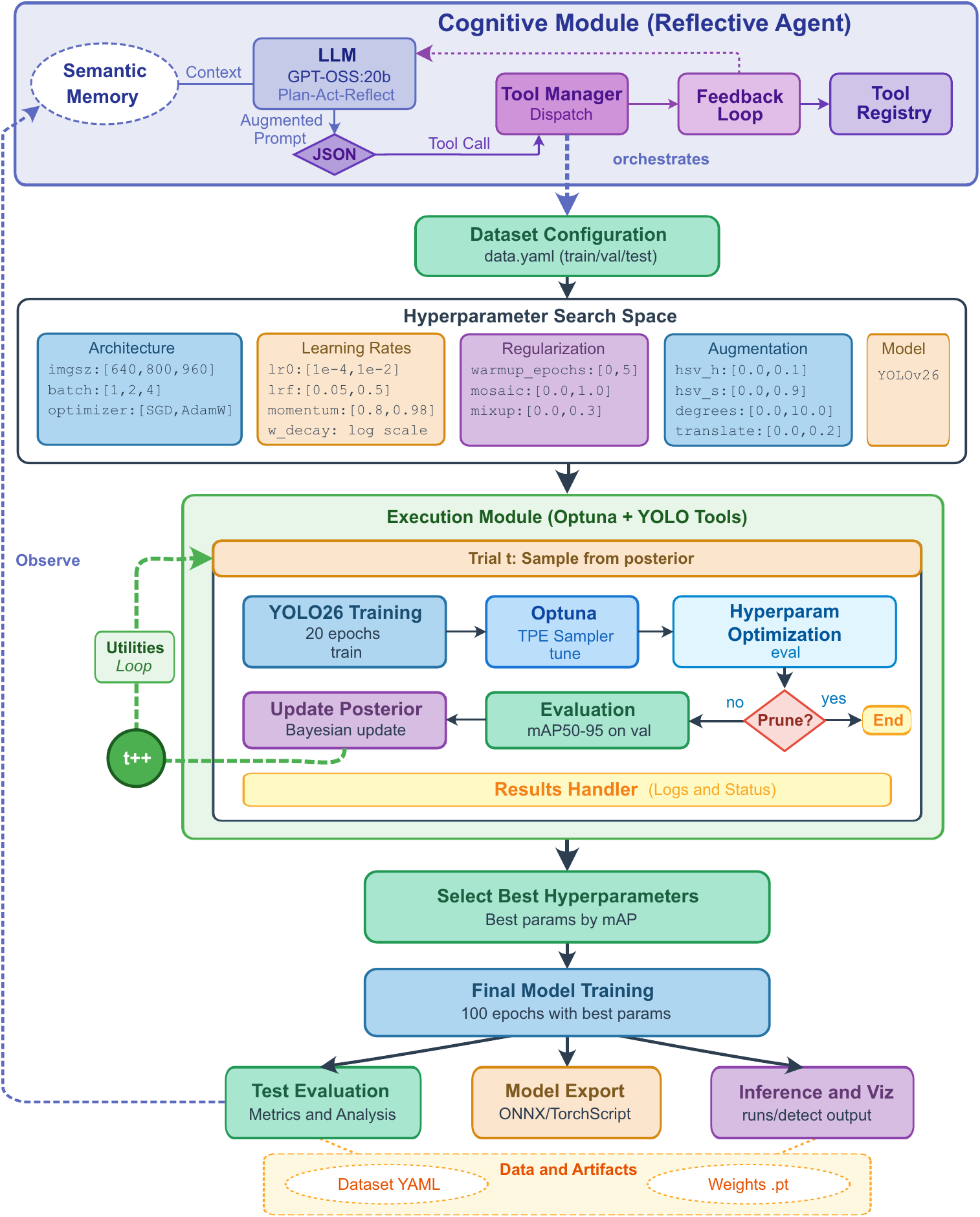}
    \caption{Workflow of the YOLO26-MoE optimization and evaluation process performed by an LLM agent.}
    \label{fig:pipeline}
\end{figure}

As illustrated in Figure~\ref{fig:pipeline}, the proposed methodology consists of a structured workflow comprising hyperparameter optimization with Optuna, final model training, evaluation on a held-out test set, and qualitative inspection through inference visualization. The LLM agent does not directly optimize model weights; instead, it orchestrates the execution of deterministic tools that implement the different stages of the pipeline. In the following subsections, we first describe the proposed YOLO26-MoE architecture and then detail the LLM-agent process.

\subsection{YOLO26-MoE}
\label{sec:yolo26_moehead}

\subsubsection{Baseline YOLO26 Detector}

YOLO26 follows the YOLO paradigm, in which object detection is formulated as a single-stage dense prediction problem that directly maps image pixels to bounding-box coordinates and class probabilities~\cite{CARVALHO2026104067}. Given an input image $I \in \mathbb{R}^{H \times W \times 3}$, the network extracts hierarchical feature representations through a lightweight convolutional backbone and predicts detections at multiple spatial resolutions.

Unlike transformer-based detectors that separate localization and classification through decoder attention layers, YOLO26 preserves a fully convolutional topology with efficient feature propagation across stages~\cite{sapkota2025YOLO26}. Its architecture is designed to maintain a favorable trade-off between computational efficiency and detection accuracy, making it suitable for edge-oriented applications. The detector operates on multiple scales, enabling the localization of objects with different apparent sizes in the image.

The original YOLO26 architecture used in this work consists of a convolutional backbone followed by a multi-scale detection head with three prediction levels: P3/8, P4/16, and P5/32. The backbone progressively downsamples the input and extracts increasingly semantic feature maps, while the head upsamples and fuses intermediate representations to recover spatial detail for small- and medium-scale objects. The final Detect module receives three feature maps and produces bounding-box and class predictions at the three scales.

\subsubsection{Proposed MoE-based Head Modification}

Although the baseline YOLO26 already provides an efficient multi-scale detection architecture, its feature refinement blocks are static, meaning that the same convolutional transformation is applied to all input samples regardless of scene content or defect morphology. In the present work, we modify the small-object detection branch of YOLO26 by replacing the original P3/8 feature refinement block with a sparse MoE module. This yields the proposed YOLO26-MoE architecture.

\begin{figure}[!ht]
    \centering
    \includegraphics[width=0.61\linewidth]{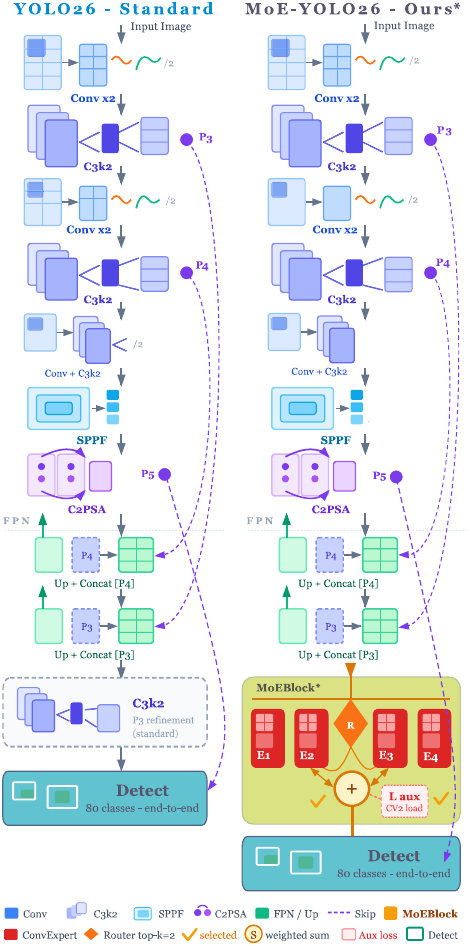}
    \caption{Comparison between the standard YOLO26 architecture and the proposed YOLO26-MoE variant. 
    }
    \label{fig:moe_yolo_architecture}
\end{figure}

Figure~\ref{fig:moe_yolo_architecture} illustrates the architectural difference between the standard YOLO26 detector and the proposed YOLO26-MoE. As can be observed, the modification is deliberately localized to the high-resolution P3 branch, where the original refinement block is replaced by a sparse MoEBlock. This design enables conditional feature refinement through routed expert selection while preserving the remaining backbone, neck, and multi-scale detection structure of the baseline architecture.

More specifically, in the original YOLO26 head, the P3 branch is formed after upsampling and concatenating higher-level features with the backbone P3 features, followed by a C3k2 refinement block. In the proposed architecture, this C3k2 block is replaced by a custom MoEBlock, while the P4 and P5 branches remain unchanged. Therefore, the architectural modification is localized to the highest-resolution prediction path, which is the branch most directly associated with small-object representation.

Let $F_{P3}\in\mathbb{R}^{B\times C\times H\times W}$ denote the input feature tensor of the modified P3 branch after concatenation. Instead of processing $F_{P3}$ with a single shared convolutional block, the proposed MoE module first projects the input to an internal representation and then routes each sample in the mini-batch to a subset of specialized experts. The output of the MoE block can be written as
\begin{equation}
Y_{P3} = \sum_{j=1}^{K} \alpha_j \, E_{\pi_j}(F_{P3}),
\end{equation}
where $E_{\pi_j}(\cdot)$ denotes the $\pi_j$-th selected expert, $\alpha_j$ is its routing weight, and $K$ is the number of active experts selected for each sample through top-$K$ routing.

The routing decision is produced by a lightweight gating subnetwork. Given the input feature tensor $F_{P3}$, global average pooling is first applied to obtain a compact channel descriptor, which is then processed by a small multilayer perceptron to generate routing logits:
\begin{equation}
\mathbf{z} = R(F_{P3}) \in \mathbb{R}^{E},
\end{equation}
where $E$ is the total number of experts. Routing probabilities are then obtained through a softmax operation:
\begin{equation}
\mathbf{p} = \mathrm{softmax}(\mathbf{z}),
\end{equation}
and the top-$K$ experts are selected according to the largest logits in $\mathbf{z}$. Their normalized routing weights are computed as
\begin{equation}
\alpha_j = \frac{\exp(z_{\pi_j})}{\sum_{m=1}^{K}\exp(z_{\pi_m})}, \qquad j=1,\dots,K.
\end{equation}

Each expert is implemented as a lightweight convolutional subnetwork composed of a spatial convolution, batch normalization, SiLU activation, and a pointwise projection back to the target channel dimension. This design allows the expert pool to specialize in different local appearance patterns while keeping the computational overhead moderate through sparse activation. In the proposed implementation, only the selected experts are evaluated for each input sample, which preserves conditional computation and avoids the cost of dense expert aggregation.

The motivation for inserting the MoE module specifically into the P3 branch is related to the characteristics of the target application. In insulator defect detection, damaged regions such as fractures and flashover traces frequently occupy limited spatial regions and may present high intra-class variability, subtle texture changes, and ambiguous boundaries. Since the P3 branch operates at higher spatial resolution, it is particularly relevant for preserving fine structural information. By replacing the static refinement block at this stage with a sparse MoE module, the detector gains the ability to adapt feature processing according to the visual characteristics of each sample, potentially improving discrimination of subtle and heterogeneous defect patterns.

\subsubsection{Detection Head and Prediction Process}

After feature refinement, the YOLO26-MoE detector preserves the original multi-scale prediction mechanism of YOLO26. Detection is performed at three scales corresponding to the P3, P4, and P5 feature maps. For each feature tensor $F_l$ at level $l$, the detection head predicts objectness, bounding-box regression values, and class scores at each spatial location:
\begin{align}
p_{obj}^{(i,j)} &= \sigma(f_{obj}(F_{l}^{(i,j)})), \\
b^{(i,j)} &= g(F_{l}^{(i,j)}), \\
c^{(i,j)} &= \mathrm{softmax}(h(F_{l}^{(i,j)})),
\end{align}
where $p_{obj}^{(i,j)}$ denotes the objectness probability, $b^{(i,j)}=(x,y,w,h)$ denotes the predicted bounding-box parameters, and $c^{(i,j)}$ is the class probability vector. The functions $f_{obj}$, $g$, and $h$ are lightweight learned mappings implemented by the detection head. Because the architectural change is restricted to the P3 feature refinement block, the overall detection logic and output structure remain compatible with the baseline YOLO26 framework.

\subsubsection{Auxiliary Expert-balancing Loss and Training Objective}

A common issue in MoE training is routing collapse, in which only a small subset of experts is repeatedly selected while the remaining experts receive little or no training signal~\cite{fedus2022switchtransformersscalingtrillion}. To mitigate this problem, the proposed MoE block includes an auxiliary balancing loss that promotes more uniform expert utilization.

Let $\mathbf{p}\in\mathbb{R}^{E}$ denote the average routing importance across a mini-batch, computed from the softmax probabilities produced by the router, and let $\mathbf{l}\in\mathbb{R}^{E}$ denote the empirical expert selection frequency across the same batch. The auxiliary MoE regularization term is defined as
\begin{equation}
\mathcal{L}_{aux} = \mathrm{CV}^2(\mathbf{p}) + \mathrm{CV}^2(\mathbf{l}),
\end{equation}
where $\mathrm{CV}^2(\cdot)$ denotes the squared coefficient of variation. Minimizing this term encourages a more balanced distribution of routing importance and actual expert usage, while still allowing experts to specialize.

The final training objective of the proposed detector is defined as
\begin{equation}
\mathcal{L}_{total} = \mathcal{L}_{YOLO26} + \lambda(t)\mathcal{L}_{aux},
\end{equation}
where $\mathcal{L}_{YOLO26}$ is the original YOLO26 detection loss and $\lambda(t)$ is a progressive weighting factor controlling the contribution of the MoE balancing loss.

Rather than applying the full MoE regularization strength from the beginning of training, a linear warmup strategy is adopted:
\begin{equation}
\lambda(t)=\lambda_{0}\min\left(1,\frac{t}{T_w}\right),
\end{equation}
where $\lambda_{0}$ is the target auxiliary-loss weight and $T_w$ is the number of warmup iterations. This design prevents the auxiliary routing constraint from dominating the early optimization dynamics, when both the detector and the routing network are still unstable. As training progresses, the balancing term is gradually strengthened, encouraging more stable expert utilization without disrupting early representation learning.

In practical terms, this behavior is implemented by extending the original Ultralytics loss routine. Before each forward pass, the MoE auxiliary collector is cleared to ensure that only losses generated in the current iteration are considered. After the standard YOLO26 detection loss is computed, the differentiable auxiliary losses produced by the active MoE blocks are summed and added to the scalar training loss before backpropagation, yielding a joint optimization procedure for detection performance and expert balancing.

\begin{figure}[!ht]
    \centering
    \includegraphics[width=0.5\linewidth]{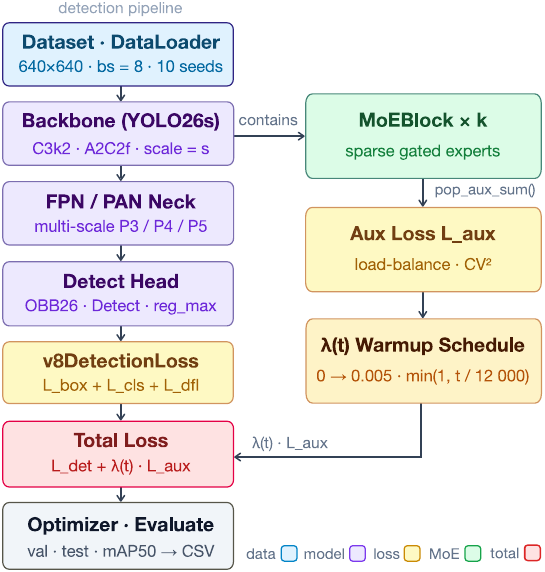}
    \caption{Training pipeline of the proposed YOLO26-MoE detector. The workflow combines standard YOLO26 detection loss with the auxiliary MoE balancing loss, weighted by a progressive warmup schedule, and jointly optimized during training.}
    \label{fig:yolo26s_moe_pipeline}
\end{figure}

Figure~\ref{fig:yolo26s_moe_pipeline} summarizes the overall training pipeline of the proposed detector, including the interaction between the backbone-neck feature extraction stages, the MoE-enhanced P3 branch, the auxiliary expert-balancing term, and the final joint optimization procedure.

\subsubsection{Deployment Considerations}

The proposed YOLO26-MoE preserves the deployment compatibility of the Ultralytics framework~\cite{sapkota2025ultralytics}. Since the modification is localized to the P3 branch and implemented using standard PyTorch modules, the architecture remains compatible with common optimization and inference pipelines. Moreover, because the MoE block operates only during feature refinement and does not alter the detector output interface, model evaluation, visualization, and export remain integrated with the standard Ultralytics toolchain. Thus, YOLO26-MoE extends the baseline YOLO26 detector with conditional computation in the high-resolution detection branch. The main objective of this modification is to improve feature adaptability for fine-grained insulator defect patterns while maintaining a lightweight and deployment-oriented detection framework.

\subsection{LLM Agent}
\label{sec:llm_agent}

A tool-augmented LLM agent is employed to standardize and automate the hypertuning, training, evaluation, and qualitative inspection of YOLO26-MoE for multi-class insulator condition detection, namely \texttt{no\_issues}, \texttt{broken}, and \texttt{flashover\_damage}. The agent does not update network parameters directly. Instead, it performs constrained decision-making over a fixed set of executable tools that deterministically implement training and evaluation procedures. This separation between LLM-based control and tool-based computation reduces manual hyperparameter selection, improves experimental repeatability, and provides traceable provenance through structured logs and persisted optimization studies.

\subsubsection{Action Model and Tool Interface}

At each interaction step $t$, the agent outputs a single action $a_t$ represented by a tool invocation:
\begin{equation}
a_t = (\tau_t, \mathbf{x}_t), \quad \tau_t \in \mathcal{T}, \;\mathbf{x}_t \in \mathcal{X}_{\tau_t},
\end{equation}
where $\tau_t$ is a tool name from a finite registry $\mathcal{T}$ and $\mathbf{x}_t$ is a typed argument dictionary. The LLM output is constrained to a single \texttt{JSON} object encoding $(\tau_t,\mathbf{x}_t)$, which is parsed and executed without manual intervention. The canonical schema is:

\begin{verbatim}
{
  "tool_call": {
    "name": "<tool_name>",
    "arguments": { "<arg1>": "<value1>", ... }
  }
}
\end{verbatim}

This strict action schema prevents free-form responses from affecting execution and ensures that each decision is accountable. Tool outputs, including metrics and artifact paths, are returned as \texttt{JSON} observations and may be re-injected into the agent context.

\subsubsection{Tool Registry}
\label{sec:tool_registry}

The tool registry $\mathcal{T}$ is designed to mirror the experimental protocol and to provide a minimal and deterministic interface between the LLM policy and the training/evaluation backend. Each tool encapsulates one pipeline stage and returns a structured \texttt{JSON} payload containing primary performance indicators, execution metadata, and artifact locations such as checkpoints, plots, and annotated predictions.

The \texttt{yolo\_optuna\_tune} tool performs Optuna-based hyperparameter optimization with the objective of maximizing validation mAP@[0.5:0.95]. The search space spans training hyperparameters, optimization parameters, and data augmentation magnitudes. For each trial, training is executed under a fixed budget of epochs, followed by validation on the validation subset. The returned \texttt{JSON} includes the best objective value and the best hyperparameter configuration $\lambda^{*}$.

The \texttt{yolo\_train\_final} tool executes the final training schedule using $\lambda^{*}$. The selected hyperparameters are merged into a single configuration and applied for an extended number of epochs with standard convergence controls such as cosine learning-rate scheduling and patience-based stopping.

The \texttt{yolo\_eval\_test} tool evaluates a specified checkpoint on the held-out test dataset to obtain an unbiased estimate of generalization performance. Evaluation is performed with fixed confidence and IoU thresholds and returns mAP@[0.5] and mAP@[0.5:0.95]. When available, the confusion matrix is also returned in the \texttt{JSON} payload, enabling class-level error analysis.

The \texttt{yolo\_infer\_visualize} tool performs inference on representative samples and stores annotated outputs for qualitative inspection. The returned \texttt{JSON} provides the output directory and an explicit list of generated artifacts, supporting transparent inspection of localization quality, missed detections, and recurrent failure patterns.

\subsubsection{Prompt Specification and Output Contract}
\label{sec:prompt_contract}

Agent behavior is controlled through a system prompt that formalizes an execution contract between the LLM policy and the tool runtime. The prompt constrains the agent along three main dimensions. First, it defines the task scope and domain role as object detection for power-grid insulator defects. Second, it fixes the class taxonomy, namely 0: \texttt{no\_issues}, 1: \texttt{broken}, and 2: \texttt{flashover\_damage}, ensuring semantic consistency across tuning, training, evaluation, and qualitative inspection. Third, it enforces a strict output format in which each turn must produce exactly one \texttt{JSON} tool invocation that can be deterministically parsed and executed without manual interpretation.

In addition to the interface constraints, the prompt also encodes a protocol-aligned decision policy: prioritize hyperparameter tuning to maximize validation mAP@[0.5:0.95], then train a final model with the selected configuration, evaluate on the held-out test split, and finally generate qualitative visualizations. To reduce ambiguity in tool selection, the prompt is augmented at runtime with an explicit catalog of available tools, including names, argument signatures, and short descriptions. This effectively restricts the action space to valid experimental operations.

\subsubsection{Memory Mechanism and Semantic Retrieval}
\label{sec:memory_retrieval}

The agent stores an interaction history containing user instructions, the corresponding LLM tool-call decisions, and the resulting tool observations. This record preserves the chain of decisions and outcomes required to reconstruct the optimization process, including tuned hyperparameters, selected checkpoints, evaluation thresholds, and generated qualitative artifacts.
When the interaction history becomes long, directly appending all prior messages to the prompt may be inefficient and may exceed the available context window. 

To address this limitation, an optional semantic retrieval mechanism constructs a compact task-relevant context for each new instruction. The retrieval stage embeds both the current instruction and the stored messages using a sentence embedding model and ranks previous messages by cosine similarity. The top-$k$ most relevant messages are then selected and placed in chronological order before being injected into the prompt together with the system contract. This yields a shorter context that preserves high-salience information while controlling prompt length.

Semantic retrieval affects only which historical information is presented to the LLM for action selection. Training, validation, and inference remain fully determined by deterministic tool execution, ensuring that reported performance metrics are not altered by the retrieval process.

\subsubsection{LLM Hypertuning and Evaluation Protocol}
\label{sec:llm_hpo_protocol}

Let $\lambda\in\Lambda$ denote the training and augmentation configuration explored during hypertuning. The goal is to select the configuration that maximizes validation performance:
\begin{equation}
\lambda^{*}=\arg\max_{\lambda\in\Lambda} \mathrm{mAP}_{50:95}^{\mathrm{val}}(\lambda),
\end{equation}
where $\mathrm{mAP}_{50:95}^{\mathrm{val}}$ is computed on the validation dataset. The search is implemented through Optuna, while the LLM agent is responsible for initiating the tuning stage, parameterizing the tool calls, and propagating the best configuration to subsequent stages.

Each Optuna trial samples $\lambda$ from a predefined search space covering input resolution, batch size, optimizer type, learning-rate parameters, momentum, weight decay, warmup epochs, and augmentation magnitudes. For each sampled configuration, YOLO26-MoE is trained for a fixed number of epochs and then evaluated on the validation set. To improve computational efficiency, intermediate validation performance is reported after each epoch through a callback mechanism, enabling the early termination of underperforming trials.

After hypertuning, the final model is trained for up to 100 epochs using $\lambda^{*}$, with a cosine learning-rate schedule and patience-based stopping to stabilize convergence. The best model obtained during this stage is then evaluated on the held-out test dataset using fixed confidence and IoU thresholds. These results constitute the quantitative performance measures reported in the Results section. In addition to aggregate metrics, qualitative inspection is performed by running inference on representative samples and storing annotated outputs, enabling visual assessment of localization quality and class confusions.
Algorithm~\ref{alg:llm_agent} summarizes the overall procedure.

\begin{table}[!ht]
\centering
\caption{Algorithm 1. LLM-agent  of YOLO26-MoE hypertuning and evaluation.}
\label{alg:llm_agent}
\begin{tabular}{p{0.96\linewidth}}
\toprule
\textbf{Input:} dataset YAML $D$, device $d$, number of trials $N$, visualization source $S$ (image/video). \\
\textbf{Output:} best configuration $\boldsymbol{\lambda}^{*}$, best checkpoint $w^{*}$, test metrics $\mathcal{M}_{test}$, qualitative artifacts $\mathcal{A}$. \\
\midrule
\textbf{1. Initialize} tool registry $\mathcal{T}$ (tune/train/eval/visualize/export) and interaction memory $\mathcal{H}$. \\
\textbf{2. Build} a system prompt encoding the task definition, label space, strict \texttt{JSON} tool-call schema, and tool catalog. \\
\textbf{3. Hypertune:} issue a tuning instruction and construct context from the system prompt and retrieved history (semantic retrieval optional). \\
\quad \textbf{3.1} Query the LLM to obtain a \texttt{JSON} tool invocation. \\
\quad \textbf{3.2} Execute \texttt{yolo\_optuna\_tune}; record the best objective value and the best configuration $\boldsymbol{\lambda}^{*}$. \\
\textbf{4. Final training:} execute \texttt{yolo\_train\_final} using $\boldsymbol{\lambda}^{*}$; obtain the best checkpoint $w^{*}$ (\texttt{best.pt}). \\
\textbf{5. Test evaluation:} execute \texttt{yolo\_eval\_test} with $w^{*}$; obtain $\mathcal{M}_{test}$. \\
\textbf{6. Qualitative inspection:} execute \texttt{yolo\_infer\_visualize} with $w^{*}$ on $S$; obtain artifacts $\mathcal{A}$ under \texttt{runs/detect}. \\
\bottomrule
\end{tabular}
\end{table}

\section{Results and Discussion}
\label{sec5}

In this section, the results of applying the proposed method are presented and discussed. First, we present the experimental setup considered to compute the experiments; after that, we present the results of applying the proposed methodology and a comparative analysis (benchmarking) to other models.

\subsection{Experimental Setup}

The experimental setup defines the hardware, dataset, and evaluated measures used in this section, making it possible to compare our results with those of other models and to facilitate future comparisons.

\subsubsection{Hardware and Software Specification}
All experiments were conducted using a standardized software and hardware environment to ensure reproducibility and fair comparison of results. The object detection framework was based on \texttt{Ultralytics} version 8.4.6, implemented in Python 3.9.25. Model training and inference were performed using \texttt{PyTorch} version 2.8.0 with CUDA support. All computations were accelerated on a single NVIDIA Quadro RTX 5000 GPU, equipped with 16 GB of dedicated memory, which provided sufficient computational capacity for efficient training and evaluation of the evaluated models.

\subsubsection{Dataset Description}

The dataset employed in the analysis presented in this paper consists of images of insulator strings as the primary class, with three subclasses representing the condition of the insulator shell: flashover-damaged insulator shell, broken insulator shell, and intact insulator shell. The data comprise original high-resolution images acquired during inspections of high-voltage transmission lines. For reproducibility and future comparative studies, the original dataset, along with detailed documentation, is publicly available in the repository created by Lewis and Kulkarni \cite{data}. 

\begin{figure}[h!]
\subfloat[]{\includegraphics[width=0.5\columnwidth]{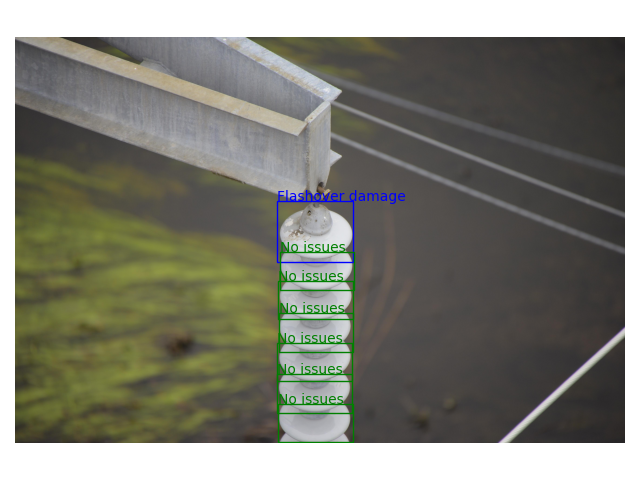}}%
\hfil
\subfloat[]{\includegraphics[width=0.5\columnwidth]{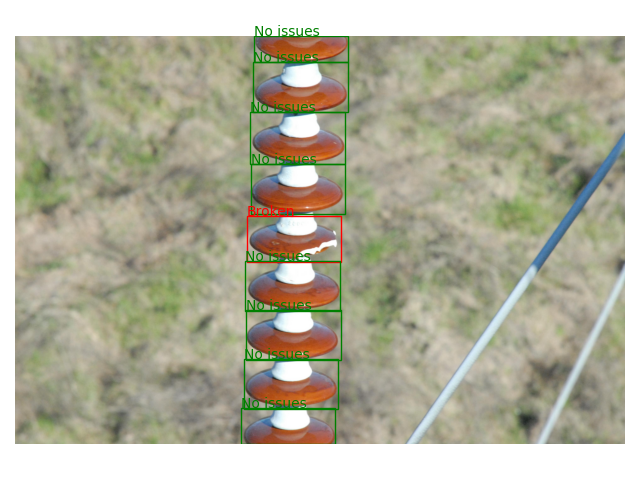}}%
\caption{Samples from the original dataset highlighted: a) flashover and b) broken insulators.}
\label{fig:data_i}
\end{figure}

High-resolution pictures taken by digital single-lens reflex cameras during power grid inspections in favorable weather circumstances make up the dataset under consideration.  The pre-processing involved rescaling the photos to 640$\times$640 pixels, which is the usual size for the models under consideration, and converting the annotations from \texttt{JSON} files to YOLO-compatible readable files (\texttt{.txt}). An example of the considered insulators (highlighting the damaged ones in a blue box and broken insulators represented by a red box) is shown in Figures \ref{fig:data_i} a) and b).

\subsubsection{Considered Measures}

The results for precision (\ref{pr}), recall (\ref{re}), F1-score (\ref{f1}), and mAP (\ref{map}) are reported. In particular, the evaluation includes mAP@[0.5] and mAP computed over an Intersection over Union threshold ranging from 0.5 to 0.95 (mAP@[0.5:0.95]). All these performance metrics are derived from the true positives ($tp$), false positives ($fp$), and false negatives ($fn$) in the classification task, and are calculated as follows:

\begin{equation}
\textup{precision}=\frac{tp}{tp+fp}
\label{pr}
\end{equation}

\begin{equation}
\textup{recall}=\frac{tp}{tp+fn}
\label{re}
\end{equation}

\begin{equation}
\textup{F1-score}=\frac {2 \times \textup{recall} \times \textup{precision}}{\textup{recall}+\textup{precision}}
\label{f1}
\end{equation}
\color{black}
\begin{equation}
\textup{mAP}=\frac{1}{n}\sum_{k=1}^{n}\left ({\sum_{\eta}(\textup{recall}_\eta-\textup{recall}_{\eta-1})\textup{precision}_\eta}\right )_k
\label{map}
\end{equation}
where $\eta$ denotes the $n$th threshold and $k$ represents the corresponding class among the $n$ classes.

\subsubsection{Compared Models}
\label{sec:compared_models}

To assess the effectiveness of the proposed YOLO26-MoE architecture, comparisons are conducted against several contemporary YOLO families, namely YOLOv10~\cite{10930899}, YOLO11~\cite{11053860}, YOLO12~\cite{11113267}, and YOLO26~\cite{sapkota2025YOLO26}, considering multiple scale variants for each architecture, including nano (n), small (s), medium (m), large (l), and extra-large (x). These models were selected because they represent successive design evolutions of the YOLO framework, ranging from efficiency-oriented lightweight detectors to higher-capacity variants with enhanced representational power. Such diversity provides a comprehensive basis for evaluating both the detection performance and computational cost of the proposed method.

For benchmarking, each YOLO family is evaluated through its scale variants in order to analyze the trade-off between accuracy and computational efficiency. The considered metrics include mAP@0.5, mAP@0.5:0.95, precision, recall, F1-score, training time, and validation time. This experimental design enables a comparative analysis not only across different YOLO generations but also across different model capacities within the same family.

In addition to the baseline detectors, the proposed YOLO26-MoE model is included as the final comparative configuration. This model extends the original YOLO26 architecture by incorporating a sparse MoE module in the high-resolution detection branch and by adopting an LLM-guided hyperparameter optimization strategy. Therefore, the benchmark is designed to verify whether the proposed architectural modification and optimization pipeline can provide measurable gains over standard YOLO baselines in the context of UAV-based insulator fault detection.

\subsection{Hypertuning Study}
\label{sec:hypertuning_study}

To identify a strong training configuration for the proposed YOLO26-MoE detector model, a hyperparameter optimization study was conducted through the tool-augmented LLM pipeline introduced in Section~\ref{sec:llm_hpo_protocol}. The LLM agent invokes a deterministic Optuna-based tool that performs the search automatically over a predefined hyperparameter space. The optimization objective is the validation mAP@[0.5:0.95], which is used as the scalar criterion for ranking candidate configurations.

The hypertuning procedure was executed over 50 Optuna trials. In each trial, the proposed model configuration was instantiated and trained for 20 epochs on the training split, followed by validation on the \texttt{val} split. During training, the validation metric \texttt{mAP50-95(B)} was reported to Optuna at the end of each epoch through a callback mechanism, allowing underperforming trials to be pruned whenever appropriate. After training, the final validation metrics were computed, and the resulting mAP@[0.5:0.95] was returned as the trial objective value.

The search space jointly covered optimization, regularization, and augmentation parameters. More specifically, the explored hyperparameters were the input image size (\texttt{imgsz}), batch size (\texttt{batch}), optimizer type, initial learning rate (\texttt{lr0}), final learning-rate factor (\texttt{lrf}), momentum, weight decay, warmup duration, and several augmentation parameters, namely hue, saturation, and value perturbations (\texttt{hsv\_h}, \texttt{hsv\_s}, \texttt{hsv\_v}), as well as mosaic, mixup, rotation, and translation. The best configuration obtained from the study is summarized in Table~\ref{tab:best_hyperparameters}.

\begin{table}[!ht]
\centering
\caption{Best hyperparameter configuration obtained for YOLO26-MoE during the Optuna-based hypertuning study.}
\label{tab:best_hyperparameters}
\begin{tabular}{lc}
\toprule
Hyperparameter & Best value \\
\midrule
Input size (\texttt{imgsz}) & 960 \\
Batch size (\texttt{batch}) & 10 \\
Optimizer & AdamW \\
Initial learning rate (\texttt{lr0}) & 0.00108 \\
Final LR factor (\texttt{lrf}) & 0.05947 \\
Momentum & 0.90001 \\
Weight decay & 0.000282 \\
Warmup epochs & 3.487 \\
Hue augmentation (\texttt{hsv\_h}) & 0.05574 \\
Saturation augmentation (\texttt{hsv\_s}) & 0.44839 \\
Value augmentation (\texttt{hsv\_v}) & 0.69333 \\
Mosaic & 0.45098 \\
Mixup & 0.20653 \\
Rotation (\texttt{degrees}) & 1.64896 \\
Translation & 0.18933 \\
\bottomrule
\end{tabular}
\end{table}

As shown in Table~\ref{tab:best_hyperparameters} the choice of a high input resolution indicates that the proposed detector benefits from preserving fine spatial detail, which is consistent with the small and localized nature of broken and flashover-damaged insulator regions. The selection of AdamW further suggests that decoupled regularization contributed to a more favorable optimization trajectory.

\begin{figure}[!ht]
    \centering
    \includegraphics[width=1\linewidth]{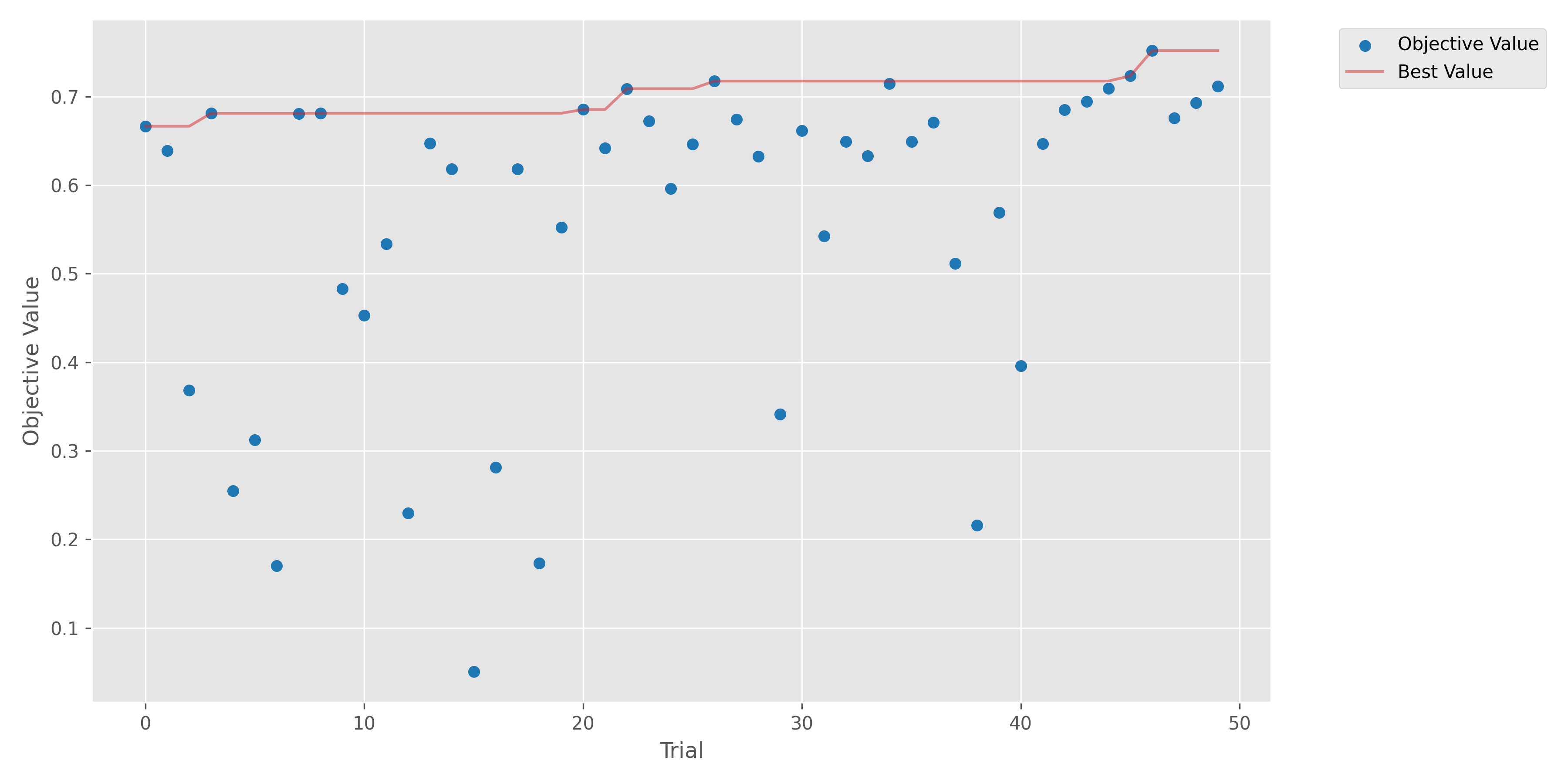}
    \caption{Optimization history of the Optuna study over 50 trials. Blue markers denote the objective value obtained in each trial, while the red curve indicates the best value achieved up to that trial.}
    \label{fig:optimization_history}
\end{figure}

As shown in Figure~\ref{fig:optimization_history}, the optimization history exhibits a progressive improvement of the objective value over the 50 trials. Although several low-performing configurations were sampled during the search, the best-so-far curve increased steadily, indicating that the Optuna study refined the explored configurations and converged toward stronger solutions.

\begin{figure}[!ht]
    \centering
    \includegraphics[width=1\linewidth]{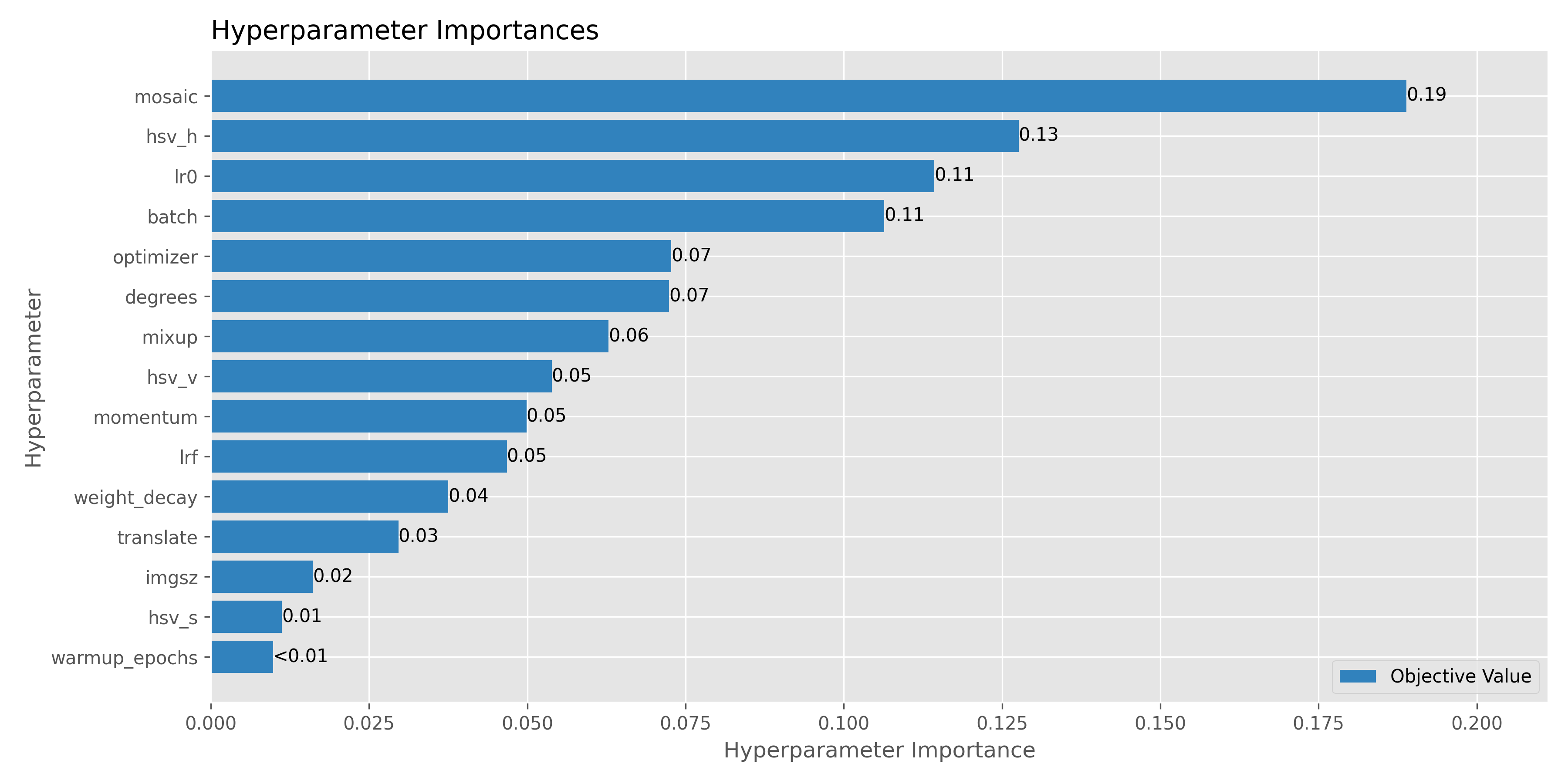}
    \caption{Relative hyperparameter importance with respect to the validation objective in the Optuna-based hypertuning study. Larger values indicate a stronger contribution to the achieved performance.}
    \label{fig:hyperparameter_importances}
\end{figure}

Figure~\ref{fig:hyperparameter_importances} presents the relative importance of each hyperparameter with respect to the optimization objective. The results show that \texttt{mosaic} was the most influential parameter, followed by \texttt{hsv\_h}, \texttt{lr0}, and \texttt{batch}. This indicates that, for the considered UAV-based insulator fault detection task, the final validation performance was strongly affected by both data augmentation and optimization dynamics. In particular, the prominence of mosaic augmentation suggests that robustness to changes in object scale, position, and surrounding context plays an important role in this problem. Conversely, parameters such as \texttt{warmup\_epochs}, \texttt{hsv\_s}, and \texttt{imgsz} showed a comparatively smaller individual contribution within the explored domain.
The resulting parameter set was therefore adopted for the final training phase with 500 epochs.

\begin{figure}[!ht]
    \centering
    \includegraphics[width=1\linewidth]{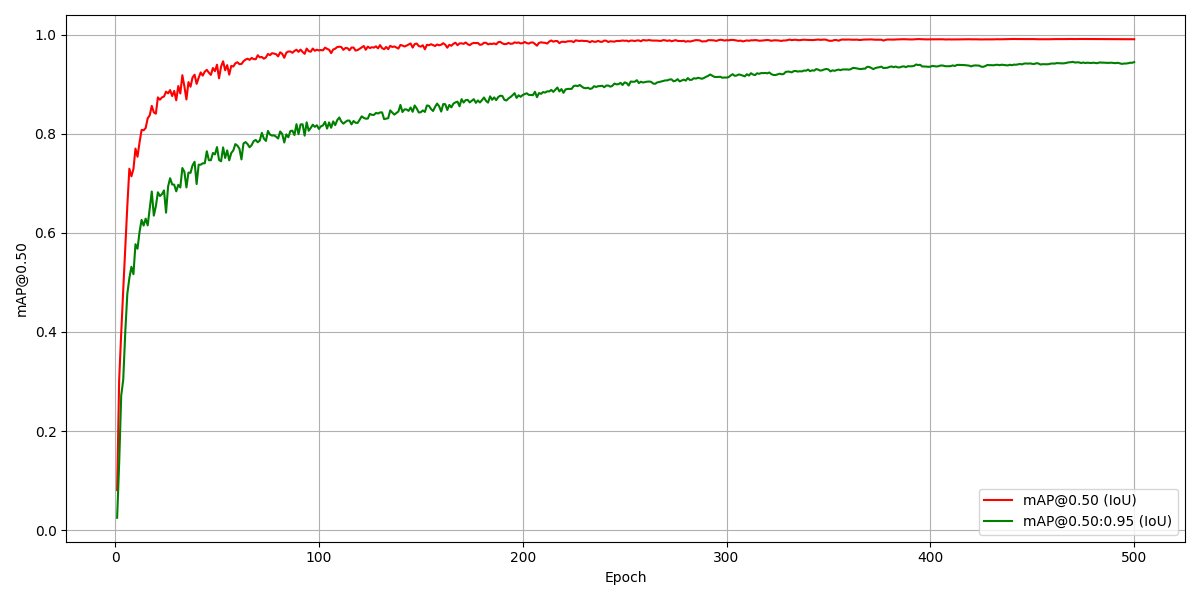}
    \caption{Evolution of mAP@0.50 and mAP@0.50:0.95 over 500 training epochs for the final optimized YOLO26-MoE configuration.}
    \label{fig:res_500_epochs}
\end{figure}

Figure~\ref{fig:res_500_epochs} shows the evolution of the main detection metrics throughout training for the final selected configuration. A rapid increase is observed during the initial epochs, followed by a smoother convergence regime, with mAP@0.50 stabilizing close to 0.99 and mAP@0.50:0.95 approaching 0.9515. This behavior indicates that the selected hyperparameter configuration not only improved the final validation objective during the Optuna search, but also yielded stable and sustained optimization during the full training schedule.

\subsection{Benchmarking Against Contemporary YOLO Models}
\label{sec:benchmarking}

To evaluate the effectiveness of the proposed detector in a broader comparative setting, benchmarking experiments were conducted against multiple recent YOLO families, namely YOLOv10, YOLO11, YOLO12, and YOLO26, considering several scale variants for each architecture. The comparison includes both detection effectiveness metrics, namely mAP@0.5, mAP@0.5:0.95, precision, recall, and F1-score, as well as computational indicators such as training time and validation time.

\begin{table}[!ht]
\centering
\caption{Performance and computational cost comparison of YOLO models}
\label{tab:YOLO_comparison_}

\begin{tabular}{lccccccc}
\toprule
Model     & mAP@0.5 & mAP@0.5:0.95 & Prec. & Recall & F1 & Train (h) & Val (s) \\
\midrule
YOLOv10n        & 0.9773  & 0.8822      & 0.9555 & 0.9255 & 0.9403 & 4.36     & 6.65   \\
YOLOv10s        & 0.9763  & 0.9099      & 0.9540 & 0.9310 & 0.9423 & 4.48     & 7.59   \\
YOLOv10m        & 0.9728  & 0.9093      & 0.9618 & 0.9330 & 0.9471 & 5.31     & 12.29  \\
YOLOv10l        & 0.9656  & 0.9079      & 0.9546 & 0.9306 & 0.9425 & 6.29     & 18.05  \\
YOLOv10x        & 0.9731  & 0.9197      & 0.9697 & 0.9346 & 0.9518 & 6.93     & 24.31  \\
\midrule
YOLO11n         & 0.9827  & 0.9047      & 0.9710 & 0.9477 & 0.9592 & 3.70     & 7.51   \\
YOLO11s         & 0.9855  & 0.9255      & 0.9794 & 0.9615 & 0.9704 & 3.86     & 8.20   \\
YOLO11m         & 0.9803  & 0.9273      & 0.9678 & 0.9507 & 0.9592 & 4.94     & 17.50  \\
YOLO11l         & 0.9850  & 0.9293      & 0.9664 & 0.9659 & 0.9661 & 5.79     & 27.60  \\
YOLO11x         & 0.9801  & 0.9211      & 0.9660 & 0.9493 & 0.9576 & 7.06     & 36.33  \\
\midrule
YOLO12n         & 0.9847  & 0.9143      & 0.9647 & 0.9581 & 0.9614 & 4.40     & 7.51   \\
YOLO12s         & 0.9884  & 0.9348      & 0.9725 & 0.9688 & 0.9706 & 4.73     & 9.89   \\
YOLO12m         & 0.9861  & 0.9311      & 0.9707 & 0.9615 & 0.9661 & 5.80     & 20.03  \\
YOLO12l         & 0.9841  & 0.9253      & 0.9604 & 0.9642 & 0.9623 & 7.75     & 25.88  \\
YOLO12x         & 0.9793  & 0.9102      & 0.9660 & 0.9473 & 0.9566 & 9.93     & 46.08  \\
\midrule
YOLO26n         & 0.9838  & 0.9134      & 0.9607 & 0.9619 & 0.9613 & 4.78     & 6.47   \\
YOLO26s         & 0.9783  & 0.9360      & 0.9753 & 0.9369 & 0.9557 & 4.87     & 8.06   \\
YOLO26m         & 0.9837  & 0.9370      & 0.9725 & 0.9487 & 0.9605 & 5.40     & 16.67  \\
YOLO26l         & 0.9859  & 0.9416      & 0.9701 & 0.9635 & 0.9668 & 6.54     & 19.17  \\
YOLO26x         & 0.9852  & 0.9366      & 0.9658 & 0.9670 & 0.9664 & 7.96     & 35.19  \\
\midrule
Proposed             & 0.9900  & 0.9515      & 0.9783 & 0.9726 & 0.9745 & 9.29 & 23.17\\
\bottomrule
\end{tabular}
\end{table}

The proposed model achieved the best overall detection performance among all evaluated configurations, reaching 0.9900 in mAP@0.5, 0.9515 in mAP@0.5:0.95, 0.9783 in precision, 0.9726 in recall, and 0.9745 in F1-score. These results indicate that the proposed architectural modification, together with the adopted hypertuning strategy, was effective in improving both localization quality and classification consistency in the considered insulator fault detection task.

When compared with the YOLOv10 family, the proposed model shows a clear advantage across all detection metrics. Although YOLOv10x was the strongest configuration within that family, with 0.9197 in mAP@0.5:0.95 and 0.9518 in F1-score, it remained consistently below the proposed approach. This suggests that the gains achieved by the proposed method are not restricted to lightweight baselines, but remain evident even against stronger variants of a recent end-to-end YOLO generation.

A similar trend is observed with the YOLO11 family. Among these models, YOLO11s provided a strong F1-score of 0.9704, while YOLO11l achieved a particularly competitive recall of 0.9659. Nevertheless, the proposed model still outperformed all YOLO11 variants in every reported detection metric, indicating a more favorable balance between feature representation, defect discrimination, and bounding-box refinement.

The YOLO12 family provided the closest competitors overall. In particular, YOLO12s reached 0.9348 in mAP@0.5:0.95 and 0.9706 in F1-score, making it one of the strongest baselines in the comparison. Even so, the proposed detector maintained a consistent advantage, especially in the stricter mAP@0.5:0.95 metric, where it improved upon YOLO12s by 0.0167 in absolute terms. This result is especially relevant because mAP@0.5:0.95 is more sensitive to localization precision and therefore provides a more demanding assessment of detector quality.

The most meaningful comparison is with the YOLO26 family, since the proposed method is directly derived from that architecture. Among the standard YOLO26 variants, YOLO26l achieved the strongest mAP@0.5:0.95 value, equal to 0.9416, while YOLO26x and YOLO26l provided the highest recall and F1 values within that family. However, the proposed YOLO26-MoE still outperformed the best baseline YOLO26 configuration, improving mAP@0.5 from 0.9859 to 0.9900, mAP@0.5:0.95 from 0.9416 to 0.9515, and F1-score from 0.9668 to 0.9745. These gains confirm that the insertion of the sparse MoE module and the adopted optimization strategy produced measurable improvements over the original YOLO26 design.

From the computational perspective, the proposed model required 9.29 h of training and 23.17 seconds of validation, which is higher than several lightweight and medium-scale baselines. Therefore, the observed gains in detection quality are accompanied by an increase in computational cost. Nevertheless, the resulting cost remains within a practical range for offline training workflows and validation-oriented studies, while delivering the best overall performance. This makes the proposed method particularly attractive for accuracy-oriented inspection scenarios in which maximizing fault detection reliability is more important than minimizing training time alone.

\begin{table}[!ht]
\centering
\caption{Model complexity comparison within the YOLO26 family in terms of parameter count and GFLOPs.}
\label{tab:model_complexity}
\begin{tabular}{lcc}
\toprule
Model & Parameters (M) & GFLOPs \\
\midrule
YOLO26n & 2.370 & 5.2 \\
YOLO26s & 9.466 & 20.5 \\
YOLO26m & 20.351 & 67.9 \\
YOLO26l & 24.748 & 86.1 \\
YOLO26x & 55.637 & 193.4 \\
Proposed & 15.659 & 58.8 \\
\bottomrule
\end{tabular}
\end{table}

To further characterize the computational trade-off of the proposed approach, Table~\ref{tab:model_complexity} reports the model complexity within the YOLO26 family in terms of number of parameters and GFLOPs. The proposed YOLO26-MoE model contains 15.659 million parameters and requires 58.8 GFLOPs, placing it between the smaller YOLO26s configuration and the larger YOLO26m/l/x variants. This result is particularly relevant because it shows that the proposed detector does not achieve its performance gains by merely scaling the baseline architecture to the highest-complexity regime. Instead, it attains superior detection performance with a complexity level that remains substantially below YOLO26m, YOLO26l, and especially YOLO26x.

This observation helps contextualize the longer training and validation times reported previously. Although the proposed model is more computationally demanding than lightweight variants such as YOLO26n and YOLO26s, it remains more compact than the larger baseline detectors while still achieving the strongest overall detection performance. Therefore, the proposed architecture provides a favorable accuracy-complexity trade-off, particularly in scenarios where detection reliability is prioritized but excessively large models are undesirable.

Overall, the benchmarking results show that the proposed YOLO26-MoE establishes the strongest accuracy-performance point among the evaluated models, especially when strict localization quality and balanced precision-recall behavior are required.

\subsection{Statistical Analysis}
\label{sec:statistical_analysis}

\subsubsection{Robustness to Random Initialization}

To assess the robustness and repeatability of the final optimized YOLO26-MoE configuration, an additional statistical analysis was conducted over 50 independent runs. The objective of this analysis was to quantify the central tendency, dispersion, and distributional characteristics of the main evaluation metrics, namely mAP@0.5, mAP@0.5:0.95, precision, recall, and F1-score.

\begin{table}[!ht]
    \centering
    \caption{Statistical summary of the performance of the optimized YOLO26-MoE model over 50 runs.}
    \label{tab:statistical_summary}
    \begin{tabular}{lrrrrr}
    \toprule
     & mAP@0.5 & mAP@0.5:0.95 & Precision & Recall & F1-score \\
    \midrule
    Mean & 0.990030 & 0.951480 & 0.978290 & 0.972650 & 0.975460 \\
    Median & 0.990500 & 0.952380 & 0.979110 & 0.972230 & 0.975330 \\
    Mode & 0.985760 & 0.946920 & 0.972880 & 0.967650 & 0.971530 \\
    Range & 0.005610 & 0.006930 & 0.010640 & 0.009210 & 0.008320 \\
    Variance & 0.000000 & 0.000010 & 0.000020 & 0.000010 & 0.000010 \\
    Std. Dev. & 0.001580 & 0.002370 & 0.004140 & 0.002880 & 0.002470 \\
    25th \%ile & 0.989930 & 0.950200 & 0.974740 & 0.970840 & 0.974040 \\
    50th \%ile & 0.990500 & 0.952380 & 0.979110 & 0.972230 & 0.975330 \\
    75th \%ile & 0.990740 & 0.953230 & 0.981820 & 0.974640 & 0.976880 \\
    IQR & 0.000810 & 0.003030 & 0.007090 & 0.003800 & 0.002830 \\
    Skewness & -2.598560 & -0.972430 & -0.155720 & -0.090230 & 0.168840 \\
    Kurtosis & 7.444460 & -0.167180 & -1.975530 & -0.594890 & -0.206260 \\
    \bottomrule
    \end{tabular}
\end{table}

Table~\ref{tab:statistical_summary} summarizes the descriptive statistics obtained across the 50 runs. The mean values remained very high for all considered metrics, reaching 0.9900 for mAP@0.5, 0.9515 for mAP@0.5:0.95, 0.9783 for precision, 0.9727 for recall, and 0.9755 for F1-score. These values are closely aligned with the corresponding medians, indicating a stable central tendency and the absence of strong discrepancies between typical and average behavior. In addition, the standard deviation and interquartile range remained low across all metrics, showing that the final detector configuration exhibits limited variability over repeated executions.

The distributional shape of the results also provides useful insight. The negative skewness observed for mAP@0.5, mAP@0.5:0.95, precision, and recall indicates that most runs are concentrated near the upper end of the performance range, with only a few lower-performing outcomes. This effect is particularly pronounced for mAP@0.5, which also exhibits a higher kurtosis, suggesting a sharply concentrated distribution around a very high performance level. Overall, these statistics indicate that the proposed detector is not only accurate but also stable and repeatable.

\begin{figure}[!ht]
    \centering
    \includegraphics[width=0.95\linewidth]{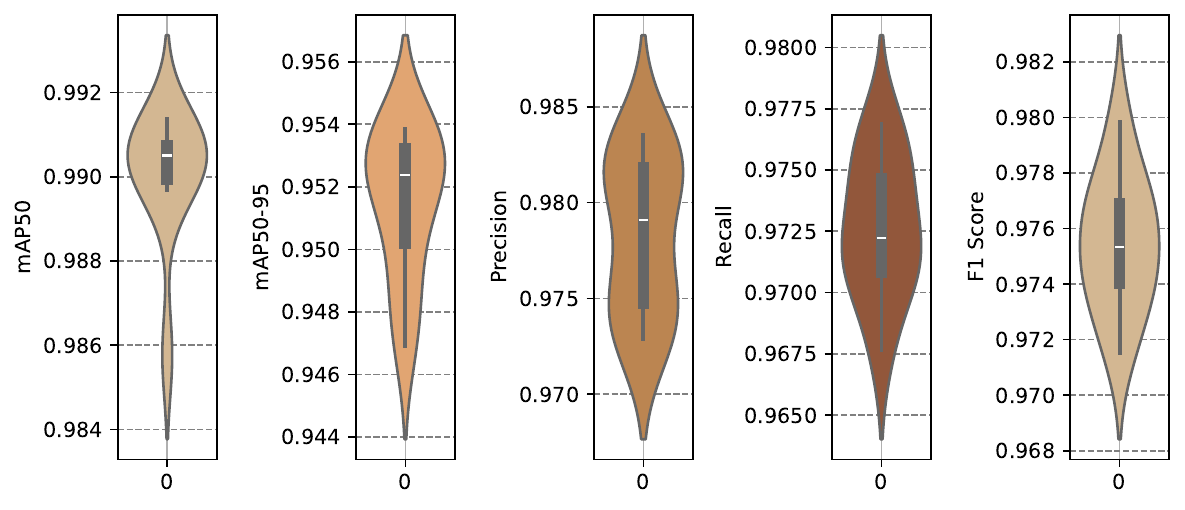}
    \caption{Violin plots of mAP@0.5, mAP@0.5:0.95, precision, recall, and F1-score over 50 runs of the optimized YOLO26-MoE model.}
    \label{fig:violinplots}
\end{figure}

These observations are visually supported by the violin plots shown in Figure~\ref{fig:violinplots}, where all five metrics present compact distributions centered around high values. In particular, the mAP@0.5 distribution is tightly concentrated around approximately 0.990-0.991, while mAP@0.5:0.95 is centered close to 0.952-0.953. Precision, recall, and F1-score also exhibit narrow and well-concentrated distributions, with no evidence of substantial multimodality or extreme spread. This visual behavior is consistent with the descriptive statistics and confirms the robustness of the final hypertuned model across repeated runs.

\subsubsection{Paired Statistical Comparison to YOLO26 Family}
\label{sec:wilcoxon_comparison}

Although the benchmarking results establish the superior overall performance of the proposed model, an additional paired inferential analysis was conducted within the YOLO26 family in order to assess whether the observed gains of the proposed YOLO26-MoE model are statistically consistent with respect to standard YOLO26 baseline variants. This analysis is particularly relevant because it evaluates the proposed detector against its closest architectural relatives, thereby isolating the contribution of the MoE modification and the adopted training strategy from broader inter-family differences.

The comparison was performed on the test split using matched random seeds. The proposed model, corresponding to the custom YOLO26-MoE configuration with warmup, was compared against the standard YOLO26 baseline variants YOLO26n, YOLO26s, YOLO26m, YOLO26l, and YOLO26x. Since the experiments were paired by seed and no normality assumption was imposed, the Wilcoxon signed-rank test was adopted as the primary inferential procedure. Holm's correction was then applied across the five pairwise comparisons within each metric. In addition, the median paired difference was computed for each comparison in order to quantify the direction and practical magnitude of the observed gains. Positive median differences indicate superior performance of the proposed YOLO26-MoE model. Due to seed overlap constraints, most paired comparisons were conducted over 10 matched runs, while the comparison against YOLO26x used 8 matched runs.

Tables~\ref{tab:wilcoxon_precision}-\ref{tab:wilcoxon_map5095} summarize the inferential comparison for precision, recall, mAP@0.5, and mAP@0.5:0.95, respectively. Overall, the proposed detector achieved statistically significant improvements over most YOLO26 baseline variants across all evaluated metrics. The only non-significant comparisons were observed against YOLO26l in precision, recall, and mAP@0.5, although even in these cases, the median paired differences remained positive. For the stricter mAP@0.5:0.95 metric, the proposed model significantly outperformed all YOLO26 baseline variants, including YOLO26l.

\begin{table}[!ht]
\centering
\caption{Paired statistical comparison between the proposed YOLO26-MoE and YOLO26 baseline variants on the test split for precision, using the Wilcoxon signed-rank test with Holm correction across the five pairwise comparisons. Positive median differences indicate superior performance of the proposed model.}
\label{tab:wilcoxon_precision}
\resizebox{\linewidth}{!}{%
\begin{tabular}{lccccc}
\toprule
Comparison & \begin{tabular}[c]{@{}c@{}}Raw \\$p$-value\end{tabular} & \begin{tabular}[c]{@{}c@{}}Holm-adjusted \\$p$-value\end{tabular} & \begin{tabular}[c]{@{}c@{}}Reject \\ $H_0$\end{tabular} & \begin{tabular}[c]{@{}c@{}}Median \\difference \end{tabular}& Direction \\
\midrule
Proposed vs YOLO26n & 0.001953 & 0.009766 & Yes & +0.008705 & Proposed $>$ YOLO26n \\
Proposed vs YOLO26s & 0.001953 & 0.009766 & Yes & +0.013132 & Proposed $>$ YOLO26s \\
Proposed vs YOLO26m & 0.009766 & 0.023438 & Yes & +0.007781 & Proposed $>$ YOLO26m \\
Proposed vs YOLO26l & 0.193359 & 0.193359 & No & +0.001583 & Proposed $>$ YOLO26l \\
Proposed vs YOLO26x & 0.007812 & 0.023438 & Yes & +0.025389 & Proposed $>$ YOLO26x \\
\bottomrule
\end{tabular}%
}
\end{table}

\begin{table}[!ht]
\centering
\caption{Paired statistical comparison between the proposed YOLO26-MoE and YOLO26 baseline variants on the test split for recall, using the Wilcoxon signed-rank test with Holm correction across the five pairwise comparisons. Positive median differences indicate superior performance of the proposed model.}
\label{tab:wilcoxon_recall}
\resizebox{\linewidth}{!}{%
\begin{tabular}{lccccc}
\toprule
Comparison & \begin{tabular}[c]{@{}c@{}}Raw \\$p$-value\end{tabular} & \begin{tabular}[c]{@{}c@{}}Holm-adjusted \\$p$-value\end{tabular} & \begin{tabular}[c]{@{}c@{}}Reject \\ $H_0$\end{tabular} & \begin{tabular}[c]{@{}c@{}}Median \\difference \end{tabular}& Direction \\
\midrule
Proposed vs YOLO26n & 0.001953 & 0.009766 & Yes & +0.012817 & Proposed $>$ YOLO26n \\
Proposed vs YOLO26s & 0.001953 & 0.009766 & Yes & +0.031801 & Proposed $>$ YOLO26s \\
Proposed vs YOLO26m & 0.001953 & 0.009766 & Yes & +0.015093 & Proposed $>$ YOLO26m \\
Proposed vs YOLO26l & 0.431641 & 0.431641 & No & +0.001183 & Proposed $>$ YOLO26l \\
Proposed vs YOLO26x & 0.007812 & 0.015625 & Yes & +0.049037 & Proposed $>$ YOLO26x \\
\bottomrule
\end{tabular}%
}
\end{table}

\begin{table}[!ht]
\centering
\caption{Paired statistical comparison between the proposed YOLO26-MoE and YOLO26 baseline variants on the test split for mAP@0.5, using the Wilcoxon signed-rank test with Holm correction across the five pairwise comparisons. Positive median differences indicate superior performance of the proposed model.}
\label{tab:wilcoxon_map50}
\resizebox{\linewidth}{!}{%
\begin{tabular}{lccccc}
\toprule
Comparison & \begin{tabular}[c]{@{}c@{}}Raw \\$p$-value\end{tabular} & \begin{tabular}[c]{@{}c@{}}Holm-adjusted \\$p$-value\end{tabular} & \begin{tabular}[c]{@{}c@{}}Reject \\ $H_0$\end{tabular} & \begin{tabular}[c]{@{}c@{}}Median \\difference \end{tabular}& Direction \\
\midrule
Proposed vs YOLO26n & 0.001953 & 0.009766 & Yes & +0.008392 & Proposed $>$ YOLO26n \\
Proposed vs YOLO26s & 0.001953 & 0.009766 & Yes & +0.016000 & Proposed $>$ YOLO26s \\
Proposed vs YOLO26m & 0.003906 & 0.011719 & Yes & +0.004639 & Proposed $>$ YOLO26m \\
Proposed vs YOLO26l & 0.083984 & 0.083984 & No & +0.001075 & Proposed $>$ YOLO26l \\
Proposed vs YOLO26x & 0.007812 & 0.015625 & Yes & +0.023990 & Proposed $>$ YOLO26x \\
\bottomrule
\end{tabular}%
}
\end{table}

\begin{table}[!ht]
\centering
\caption{Paired statistical comparison between the proposed YOLO26-MoE and YOLO26 baseline variants on the test split for mAP@0.5:0.95, using the Wilcoxon signed-rank test with Holm correction across the five pairwise comparisons. Positive median differences indicate superior performance of the proposed model.}
\label{tab:wilcoxon_map5095}
\resizebox{\linewidth}{!}{%
\begin{tabular}{lccccc}
\toprule
Comparison & \begin{tabular}[c]{@{}c@{}}Raw \\$p$-value\end{tabular} & \begin{tabular}[c]{@{}c@{}}Holm-adjusted \\$p$-value\end{tabular} & \begin{tabular}[c]{@{}c@{}}Reject \\ $H_0$\end{tabular} & \begin{tabular}[c]{@{}c@{}}Median \\difference \end{tabular}& Direction \\
\midrule
Proposed vs YOLO26n & 0.001953 & 0.009766 & Yes & +0.026397 & Proposed $>$ YOLO26n \\
Proposed vs YOLO26s & 0.001953 & 0.009766 & Yes & +0.042125 & Proposed $>$ YOLO26s \\
Proposed vs YOLO26m & 0.001953 & 0.009766 & Yes & +0.006738 & Proposed $>$ YOLO26m \\
Proposed vs YOLO26l & 0.048828 & 0.048828 & Yes & +0.001878 & Proposed $>$ YOLO26l \\
Proposed vs YOLO26x & 0.007812 & 0.015625 & Yes & +0.076509 & Proposed $>$ YOLO26x \\
\bottomrule
\end{tabular}%
}
\end{table}

The results show that the proposed detector consistently improves upon the lightweight and medium-scale YOLO26 variants, with statistically significant gains across all evaluated metrics for YOLO26n, YOLO26s, and YOLO26m. The strongest absolute improvements were generally observed against YOLO26x and YOLO26s, particularly in recall and mAP@0.5:0.95, where the median paired differences reached +0.049037 and +0.076509 against YOLO26x, and +0.031801 and +0.042125 against YOLO26s, respectively.

A particularly relevant observation concerns the comparison with YOLO26l, which was the strongest baseline in the benchmarking table. In this case, the proposed model did not show statistically significant gains in precision, recall, or mAP@0.5, although the median paired differences remained positive in all three cases. However, for mAP@0.5:0.95, the proposed model still achieved a statistically significant improvement over YOLO26l, with a Holm-adjusted $p$-value of 0.048828 and a positive median difference of +0.001878. This is especially important because mAP@0.5:0.95 is the strictest localization metric considered in this study, indicating that the proposed model improves detection quality even against the strongest baseline when more demanding IoU thresholds are taken into account.

\begin{figure}[!ht]
    \centering
    \includegraphics[width=0.5\linewidth]{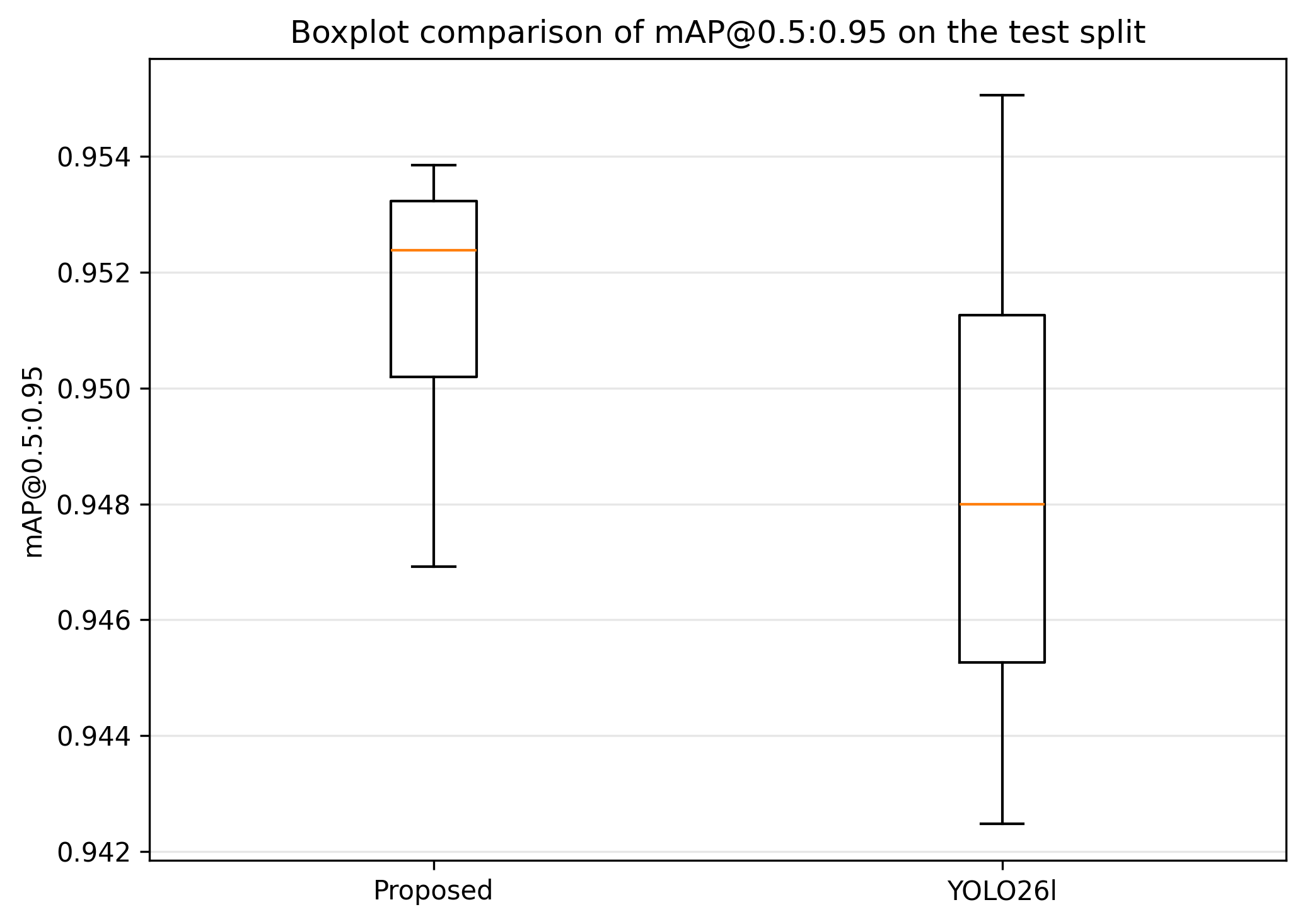}
    \caption{Boxplot comparison of mAP@0.5:0.95 on the test split between the proposed YOLO26-MoE model and the YOLO26l baseline over the matched runs.}
    \label{fig:boxplot_map5095_yolo26l}
\end{figure}

These inferential results are visually reinforced by the boxplot shown in Figure~\ref{fig:boxplot_map5095_yolo26l}, which compares the distribution of mAP@0.5:0.95 between the proposed model and the YOLO26l baseline over the matched test runs. The proposed model exhibits an upward shift in the central tendency of the distribution, together with a higher median value, which is consistent with the positive median paired difference and the statistically significant result obtained for this metric. Since YOLO26l constitutes the strongest baseline within the YOLO26 family, this visual comparison provides additional evidence that the proposed YOLO26-MoE configuration improves detection quality even against the most competitive within-family reference.

This inferential analysis complements the benchmarking and descriptive statistical results reported previously. While the benchmarking study establishes the relative position of the proposed detector among contemporary YOLO models, and the descriptive analysis demonstrates that the final configuration is stable across repeated runs, the present Wilcoxon-Holm comparison verifies that the gains observed within the YOLO26 family are statistically meaningful in most cases. Therefore, this analysis strengthens the claim that the proposed MoE modification yields systematic improvements over standard YOLO26 baseline configurations.

\subsection{Limitations}

While the proposed YOLO26-MoE model demonstrates superior performance, some limitations should be acknowledged:
\begin{itemize}
    \item \textbf{Computational Complexity:} The combination of sparse MoE processing, LLM hypertuning, and an extended optimization pipeline increases the computational cost of the proposed approach relative to lightweight YOLO26 variants. Although the final model remains less complex than larger baseline configurations such as YOLO26m, YOLO26l, and YOLO26x, it still requires higher computational resources than YOLO26n and YOLO26s, which may hinder deployment in highly resource-constrained edge environments.

    \item \textbf{Interpretability Trade-off:} While the proposed MoE design improves feature adaptability, the routing behavior of the experts adds architectural complexity that may reduce interpretability compared with the standard YOLO26 backbone.

    \item \textbf{Dependence on Search Space Design:} The effectiveness of the proposed optimization strategy depends on the predefined hyperparameter search space, the selected number of Optuna trials, and the adopted training schedule. Although the LLM agent provides a structured  layer, the final optimized configuration is still constrained by the candidate parameter ranges made available during hypertuning.

\end{itemize}

\section{Conclusion}
\label{sec6}

This paper presented YOLO26-MoE, a novel detector for UAV-based insulator fault detection that integrates a sparse MoE module into the high-resolution branch of a YOLO26 architecture and combines it with an LLM hyperparameter optimization pipeline. The proposed approach was designed to improve the adaptability of feature processing for subtle and heterogeneous defect patterns while preserving a deployment-oriented one-stage detection framework.

The experimental results demonstrated that the proposed model achieved the strongest overall performance among the evaluated detectors. In the global benchmarking analysis, YOLO26-MoE outperformed all considered YOLOv10, YOLO11, YOLO12, and YOLO26 variants, reaching the highest values of mAP@0.5, mAP@0.5:0.95, precision, recall, and F1-score. In addition, the repeated-run statistical analysis showed that the final hypertuned configuration is stable and repeatable, with high central tendency and low dispersion across 50 independent runs. The paired Wilcoxon-Holm analysis within the YOLO26 family further showed that the proposed detector yields statistically significant improvements over most baseline variants, with particularly strong gains in the stricter mAP@0.5:0.95 metric.

From the computational perspective, the proposed detector increases complexity relative to lightweight YOLO26 baselines, but remains below the larger YOLO26m, YOLO26l, and YOLO26x configurations in terms of parameter count and GFLOPs. This indicates that the observed gains are not simply the result of scaling the baseline architecture to a much larger model, but rather stem from a more effective use of model capacity through conditional expert-based feature refinement. As a result, the proposed approach offers a favorable accuracy-complexity trade-off for inspection scenarios in which detection reliability is prioritized.

\section*{Consent to Publish declaration}
Not applicable

\section*{Consent to Participate declaration}
Not applicable

\section*{Ethics declaration}
Not applicable


\section*{Acknowledgments and Funding}

The APC was supported by the project: Self-adaptive platform based on intelligent agents for the optimization and management of operational processes in logistic warehouses (PLAUTON), PID2023-151701OB-C21, funded by MCIN/AEI/10.13039/501100011033/FEDER, EU.

This work was also supported by FCT – Fundação para a Ciência e Tecnologia, I.P. under Grants 2023.15441.TENURE.051/CP00003/CT00029 and the LASIGE Research Unit (ref. UID/00408/2025, DOI: https://doi.org/10.54499/UID/00408/2025), and co-financed by FCT and the PRR - Recovery and Resilience Plan by the European Union under the LASIGE Research Unit (UID/PRR/00408/2025, \linebreak DOI: https://doi.org/10.54499/UID/PRR/00408/2025).








    \bibliographystyle{Misc/model3-num-names.bst}
    \bibliography{Misc/theBiblio}

\end{document}